\documentclass[10pt,journal,cspaper,compsoc]{IEEEtran}

\ifCLASSOPTIONcompsoc
  \usepackage[nocompress]{cite}
\else
  \usepackage{cite}
\fi

\ifCLASSINFOpdf
   \usepackage[pdftex]{graphicx}
   \graphicspath{{../pdf/}{../jpeg/}}
   \DeclareGraphicsExtensions{.pdf,.jpeg,.png,.jpg}
\else
   \usepackage[dvips]{graphicx}
   \graphicspath{{../eps/}}
   \DeclareGraphicsExtensions{.eps}
\fi

\usepackage{amsmath}
\usepackage{amsfonts}

\ifCLASSOPTIONcompsoc
  \usepackage[caption=false,font=footnotesize,labelfont=sf,textfont=sf]{subfig}
\else
  \usepackage[caption=false,font=footnotesize]{subfig}
\fi

\usepackage{stfloats}

\usepackage{amsmath}
\usepackage{amsfonts}
\usepackage{enumerate}

\def\ie{{\em i.e.}}
\def\eg{{\em e.g.}}
\def\etal{{\em et al.}}

\newcommand{\mc}[1]{\mathcal{#1}}

\newcommand{\br}[1]{\bm{\mathrm{#1}}}

\newcommand{\bs}[1]{\boldsymbol{\texttt{#1}}}

\usepackage{ragged2e}
\usepackage{booktabs}
\usepackage{color}
\usepackage{multirow}
\usepackage{bm}
\usepackage{bbm}

\usepackage{xcolor}
\usepackage{balance}

\usepackage{amssymb}

\graphicspath{{./figures/}}

\usepackage{dsfont}
\usepackage{hyperref}
\usepackage{cleveref}
\usepackage{algorithm2e}
\RestyleAlgo{ruled}

\Crefname{algocf}{alg.}{algs.}
\Crefname{algocf}{Algorithm}{Algorithms}

\begin{document}

\title{
Adapting Dense Vision-Language Relationships for Multi-label Classification with Partial Label
}

\author{Cheng Chen,~Yifan Zhao,~\IEEEmembership{Member,~IEEE}, and Jia Li,~\IEEEmembership{Senior Member,~IEEE}
\IEEEcompsocitemizethanks{
  \IEEEcompsocthanksitem C. Chen, Y. Zhao and J. Li are with the State Key Laboratory of Virtual Reality Technology and Systems, School of Computer Science and Engineering, Beihang University, Beijing, 100191, China.
  \IEEEcompsocthanksitem J. Li and Y. Zhao are the corresponding authors (E-mail: jiali@buaa.edu.cn,
  zhaoyf@buaa.edu.cn, ). 
}}

\IEEEtitleabstractindextext{%
\begin{abstract}
\justifying Learning multi-label image classification with incomplete annotations is a challenging task that has been widely studied for its superior trade-off between high efficiency and less labor consumption on large-scale datasets. Predominant methods rely on strong prior assumptions to recover the missing semantics from partial annotations. However, these statistic priors suffer from unstable semantic mistakes and thus lead to catastrophic overfitting. Toward this end, we propose a Language-driven Dense Semantic Adaptor (LDSA) that excavates prior-adaptive relationships from multimodal pretrained CLIP models. In our approach, the densely contrastive adaptor is first proposed to construct dense visual contrastive constraints, transferring the task-specific knowledge to visual domains. We then propose a language-driven interactive decoder with the help of class-specific prompt tuning, which adapts language proxies with visual domains. With the collaborative learning of proposed modules, experimental results demonstrate our proposed LDSA achieves a new state of the art on public multi-label classification benchmarks, and interpretable analyses reveal that our LDSA discovers implicit semantic relationships with the prior-adaptive learning scheme.
\end{abstract}

}

\maketitle

\IEEEdisplaynontitleabstractindextext

\IEEEpeerreviewmaketitle

\IEEEraisesectionheading{\section{Introduction}\label{sec:introduction}}

\IEEEPARstart{M}{ulti-label} image classification aims to predict multiple labels associated with objects present in one image, which is a vital and fundamental task in computer vision. Dozens of research~\cite{Guo2023textprompt, Zhu2023multiLabelSceneLearning, Zhu2023sceneGraphMultiLabel, Li2023CTMultiLabel, Zhang2022MLContrastive, gcn_tpami,sst_tip,mcra_tip,Chen2022KGGR,tdrg_ICCV2021,mlgcn_CVPR2019} have made significant progress in solving this fully-supervised problem, which can be largely attributed to the consolidation of large-scale datasets. Due to the difficulties in constructing large-scale datasets, many recent studies~\cite{role_CVPR2021,sst_aaai2022,Pu2022SARB,Rajeswar_2022_CVPR,Kim2023gapOfPartialLabel, Zhang2023learnLongTailedPartial} have concentrated their efforts on solving multi-label classification problems with only partial labels.

\begin{figure}[t]
  \centering
  \includegraphics[width=\linewidth]{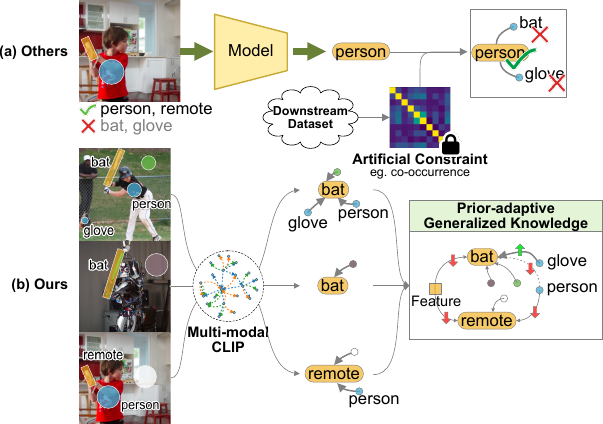}
  \caption{The motivation of our approach. a) Existing works learn semantic relationships with prior assumptions,~\eg, label co-occurrence, which leads to semantic mistakes in specific scenarios. b) Our approach aims to excavate adaptive relationships from generalized knowledge in CLIP and transfer them to downstream partial multi-label learning.}
  \label{fig:header}
\end{figure}

Due to the intrinsic nature of lacking annotation, partial label learning methods focus on recovering the missing labels from \textbf{prior assumptions}.  Following this trend, in \Cref{fig:header} a), 
existing methods~\cite{interactiveCNN_CVPR,Kundu2020exploitweakly} propose to model the semantic \textit{co-occurrence} by using the correlations of language labels,~\ie, the \textit{remote} and \textit{person} classes. 
While other works~\cite{sst_aaai2022,Pu2022SARB} suppose that the known labels should be uniformly distributed and rely on iterative sampling to model the co-occurrence between objects. 
Besides, recent approaches also exploit prior data knowledge by using the Bayesian uncertainty strategy~\cite{Durand_2019_CVPR} or training data statistics with prior models~\cite{role_CVPR2021,JIANG2018mentornet}. Despite these prior-based works having achieved impressive performance, they heavily rely on fixed distribution assumptions of training data. Here in \Cref{fig:header} b), we find these relationships among objects are usually not monotonous,~\eg, the object in one person's hand can be a \textit{bat} but also a \textit{TV remote}. Learning with fixed or statistic relationships in \Cref{fig:header} a) would severely mislead the object representation in such scenarios. Hence we argue that the \textbf{prior assumption in multi-label learning is a mixed blessing}, which strongly depends on the statistics of training data and limits the model to adapt to variant downstream scenarios.
The common solution to refine multi-object visual relationships is by introducing language-based priors. Prevailing works~\cite{interactiveCNN_CVPR,Kundu2020exploitweakly,Wu2015MLMG} 
rely on the predetermined text-wise relationships to guide the model optimization. To solve these drawbacks, we propose to construct vision-language relationships from the contrastive pretrained CLIP~\cite{clip} model. Different from previous works, CLIP represents the visual and language semantics in a highly aligned semantic space, which provides strong potential for excavating generalized prior knowledge from multi-modalities.  However, this generalized and aligned knowledge usually collapses during the downstream partial-label fine-tuning process. Thus one natural question arises: \textit{how to adapt the generalized pretrained CLIP knowledge to promote the learning of multi-label learning with partial annotations?}

In this paper, we propose a \textbf{prior-adaptive} multi-label learning approach, namely Language-driven Dense Semantic Adaptor (LDSA), to excavate adaptive relationships from pretrained multi-modal CLIPs. The approach aligns CLIP space by contrastively adapting the vision features and language-driven interactions.
Instead of the intuitive fine-tuning process on visual encoders, we deem the encoders as generalized feature extractors and propose a Densely Contrastive Adaptor to re-organize the classification decision boundaries,~\eg, restricted in the 80 classes on MS-COCO datasets. To achieve this, we first encourage the learning of dense semantic features with spatial resolutions instead of conventional pooling paradigms. We then build implicit contrastive relationships among different augmented views of the same object, thus these dense features would form natural semantic clusters through learning similar to conventional contrastive learning. During this process, the task-specific visual relations are extracted from the generalized semantic embedding. Despite its effectiveness, the individual adaptation of visual features would result in a misalignment of visual and language modality, making it challenging to learn from text guidance.

To mitigate the domain gap caused by visual adaptation, we resort to prompt tuning fashion~\cite{zhou2022coop,Rao2022denseclip} to force the language features to gain on the visual ones. The learnable class-specific prompts enable the language prototypes of each class to adapt independently in the high-dimensional feature space. With these adaptive language prototypes, we then propose a language-guided decoding process to refine these dense visual features. These adaptive language prototypes then serve as query proxies and correct the representation space by cross-modal visual-language interactions, where both image features and semantic prototypes are well distributed.

To sum up, this work presents an adaptive visual tuning and language refining framework to excavate prior-adaptive multi-label relationships from generalized knowledge. With this learning framework, our proposed method outperforms the state-of-the-art methods by over $2.8\%$ on MS-COCO~\cite{coco}, $9.3\%$ on Visual Genome \cite{krishna2017visual}, and $0.5\%$ on PASCAL VOC~\cite{pascal-voc-2007} benchmarks. Our contribution is three-fold: 
\begin{itemize}
    \item We present a Language-driven Dense Semantic Adaptor (LDSA) for multi-label learning with partial annotations, which excavates prior-adaptive generalized knowledge instead of the fixed prior assumptions in prevailing methods.
    \item We propose a densely contrastive adaptor to transfer the task-specific knowledge to visual domains and propose a language-driven interactive decoder with the help of prompt tuning, adapting language proxies with the visual domain.
    \item We conduct detailed experimental analyses on the prior-adaptive semantic embeddings for interpretable learning. Experimental results indicate that our proposed LDSA outperforms the prevailing methods by a large margin with fast convergence.
\end{itemize}

The remainder of this paper is organized as follows: \Cref{sect:related} provides the literature review about multi-label learning and \Cref{sect:approach} introduces the proposed Language-driven Dense Semantic Adaptor (LDSA) approach. The experimental results are reported in \Cref{sect:exp}, and \Cref{sec:interpre} further provides a detailed analysis and discussions of the proposed approach. \Cref{sec:conclusion} finally concludes this paper.

\section{Related Works} \label{sect:related}

\subsection{Multi-label Classification with Partial Annotations}
Multi-label classification is one of the most fundamental tasks in computer vision \cite{Tsoumakas_2009_MLOverview}, attracting much research attention in recent years \cite{Guo2023textprompt, Zhu2023multiLabelSceneLearning, Zhu2023sceneGraphMultiLabel, Li2023CTMultiLabel, Zhang2022MLContrastive,gcn_tpami,sst_tip,mcra_tip,Chen2022KGGR,tdrg_ICCV2021,mlgcn_CVPR2019}.
Guo \etal \cite{Guo2023textprompt} treat texts as images to enhance image recognition. Zhu \etal \cite{Zhu2023sceneGraphMultiLabel} and Chen \etal \cite{mlgcn_CVPR2019} learn the diverse co-occurrence of labels under variable scenes.
Chen \etal \cite{gcn_tpami} propose Graph Convolutional Network models to capture label dependencies to improve recognition performance. Chen \etal \cite{sst_tip} propose a module to capture both spatial and semantic correlations in multi-label images.
Gao \etal \cite{mcra_tip} propose a framework to recognize objects from global images to local regions following the way human beings perceive objects.
While works are continuously achieving better performance, due to, in part, better generalization brought by the increasing size of multi-label datasets such as ImageNet \cite{deng2009imagenet}, MS-COCO \cite{coco}, and PASCAL VOC \cite{pascal-voc-2007}, however, the labor and cost of creating such fully labeled large-scale datasets are gradually unaffordable. To tackle this, recent settings on incomplete annotations have attracted much attention \cite{role_CVPR2021,Pu2022SARB,sst_aaai2022,interactiveCNN_CVPR,Rajeswar_2022_CVPR,Ben-Baruch_2022_CVPR,Sun2022globallocal,PiCO,Kim2023gapOfPartialLabel, Zhang2023learnLongTailedPartial}. In the partial label setting, each image is annotated with only a few labels, leaving the rest unknown (either absent or present).
Cole \etal~\cite{role_CVPR2021} propose new loss with regularization terms to train models with partial annotations concerning the prior information.
Rajeswar \etal~\cite{Rajeswar_2022_CVPR} use iterated transfer learning to solve ambiguity biases.
Ben-Baruch \etal~\cite{Ben-Baruch_2022_CVPR} propose new loss or processes with regularization terms to train models with partial labels concerning the prior information. Chen \etal~\cite{sst_aaai2022}, Pu \etal~\cite{Pu2022SARB}, and Huynh \etal~\cite{interactiveCNN_CVPR} exploit correlations and feature similarities with annotated images to improve performance. Kim \etal \cite{Kim2023gapOfPartialLabel} study the impact of false negative labels to fix the model trained with partial annotations.
Most of these works pretrain their models on large datasets like ImageNet and exploit prior information provided by humans, to constrain and train the model with partial annotations.
First, we point out that these commonly used datasets are costly and suffer from incomplete annotations \cite{corruptimagenet_2019}. Second, they inherently assume that the distribution of partial labels is consistent on the downstream datasets, which in reality is hard to guarantee. Our proposed end-to-end approach just leverages unsupervised image-text knowledge and fast fine-tunes it to fit downstream datasets.

\subsection{Vision-language Models}
Vision-language models explore the alignment between visual and textual features to enhance the learning of both. CLIP \cite{clip} is a vision-language model trained on 400 million image-text pairs. It constructs a flexible prediction space with rich semantic structures, enabling CLIP's generalization and transfer. With the semantic knowledge by CLIP, recent works transfer it to various downstream tasks and achieve success \cite{Medhini2021clipit,Patashnik2021styleCLIP,Kim2022diffusionCLIP,Wang2022CRIS,Wang2022clipNerf,sun2022dualcoop,hu2023dualcoopfasteffectiveadaptation}. Since such semantic knowledge is more complex and refined than downstream datasets, approaches avoid fine-tuning the whole CLIP model but introduce prompt learning and structures with a few parameters to adapt CLIP. Zhou \etal~\cite{zhou2022coop} propose continuous prompts to explore the possibility of prompts in vision. Gao \etal~\cite{gao2021clipadapter} propose the CLIP-Adaptor, revealing different adaptation methods by adding an adaptor. Guo \etal~\cite{Guo2023textprompt} use additional texts as images in enhancing multi-label image recognition with CLIP. Yuan \etal~\cite{yuan2024positivelabelneedmultilabel} also propose method to address the noise introduced by partial label with positive labels only. These works are conducted on full-label multi-class recognition. Learning multi-label image recognition with partial labels itself suffers from overfitting and feature collapse due to the scarcity of annotations and constraints. In this task, the parameters need to be lightweight and carefully designed, which matches the requirements of adapting CLIP.

\begin{figure*}[t]
  \centering
  \includegraphics[width=0.95\textwidth]{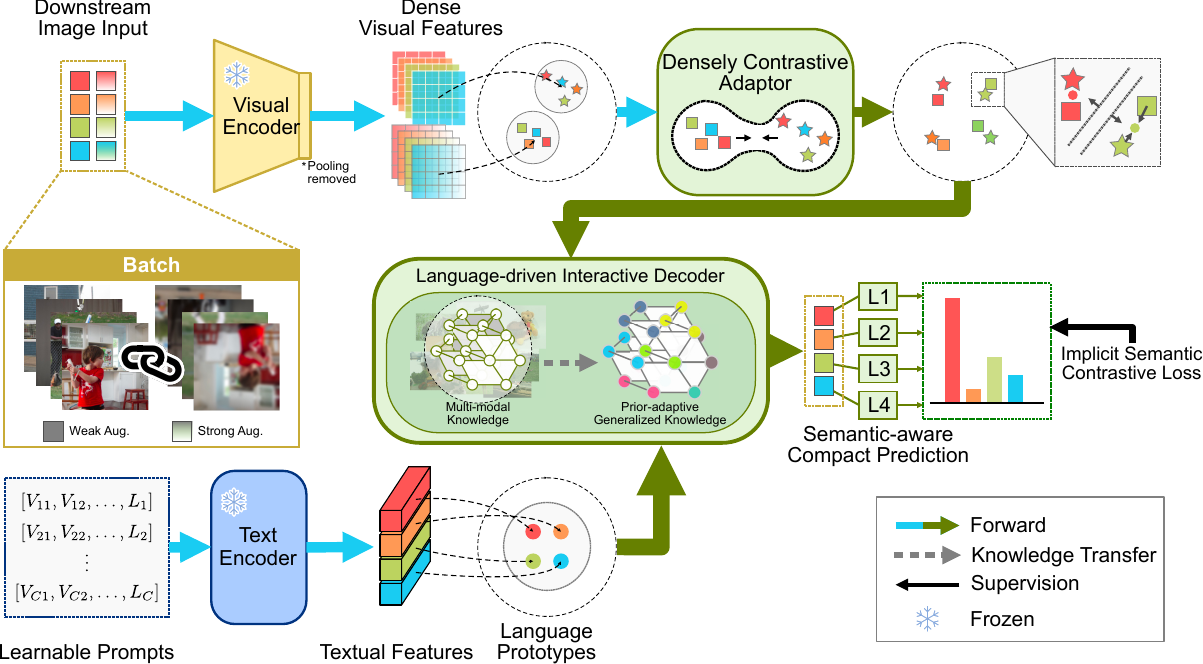}
  \caption{The proposed LDSA framework consists of two major steps: 1) The densely contrastive adaptor transfers and constrains the semantic representations from generalized CLIP visual encoders. 2) Pretrained language knowledge adapts to visual domains with prompt learning and transferring the text priors with the language-driven interactive decoder.   }
  \label{fig:pipeline}
\end{figure*}

\subsection{Contrastive Learning}
Contrastive learning (CL) is widely used in self-supervised representation learning, which aims at learning by distinguishing different instances. He \etal~\cite{moco_CVPR2020} propose a contrastive learning framework with momentum encoders and an improved sampling strategy. Grill \etal~\cite{byol} propose the BYOL model optimized with only positive samples. In addition to samplings, Chen \etal~\cite{simclr} reveal the equivalence of a memory bank and a large batch. Wang \etal~\cite{PiCO} introduce contrastive learning to partial label learning. These works perform contrastive learning on multi-class datasets, where each image contains one salient object and therefore satisfies a prerequisite of CL, \ie, semantic consistency. Multi-label datasets, however, do not satisfy semantic consistency, as images always consist of multiple objects. Radford \etal~\cite{clip} follow the contrastive learning paradigm to train the model with image-text pairs. The text (captions) can describe complex images, thus avoiding facing the semantic consistency problem. Additionally, some methods use more complex object detection to enhance object-level contrastive learning. Li \etal \cite{univip} propose a framework trying to select individual objects in multi-label images for contrastive learning. Chen \etal \cite{chen2023semantic} propose a weakly supervised segmentation bootstrapping framework to achieve instance-level contrastive learning. How to mitigate semantic inconsistency and perform contrastive learning on multi-label datasets is still under investigation.

Batch Normalization (BN)\cite{Ioffe2015batchNorm} is a technical approach like Layer Normalization \cite{Ba2016layerNorm} and Group Normalization \cite{Wu2018groupNorm}, that can accelerate the training of models. The key of BN is to normalize features across a batch, making the distribution of outputs of a layer stable for the following layers to learn.
Recent studies \cite{bnInBYOL} find that the BN can stop BYOL from feature collapse. We further exploit such a phenomenon to accomplish implicit contrastive learning on frozen CLIP.

\section{Approach} \label{sect:approach}
\subsection{Formulations and Framework}
Let $\mathcal X=\{ \mathbf x_1, \mathbf x_2, \dots, \mathbf x_N\}$ denote the training samples associated with label space $\mathcal Y=\{\mathbf y_1,\mathbf y_2,\dots,\mathbf y_N\}$. For the multi-label classification tasks, each ground truth label belongs to a binary set, ~\ie, $\br{y} \in \{-1,1\}^{C}$, where $C$ is the number of predefined semantic categories. Here $\br{y}_{i,j}=1,-1$  means the label $j$ is present (positive) or absent (negative) for the $i$-th image respectively. For multi-label learning with partial annotations, only a few labels (including positive and negative) are notated with ground truth values. Given partial label annotations $\hat{\mathcal Y}=\{\hat{\mathbf y}_1,\hat{\mathbf y}_2,\dots,\hat{\mathbf y}_N\}$, most $\hat{\br{y}}_{i,\cdot}$ are unknown for one specific image,~\ie, $\hat{\br{y}} \in \{-1,0,1\}^{C}$. $\br{y}_{i,j} = 0$ indicates whether the $j$th category present in $i$th image is unknown.

In this learning setting, only partial labels $\hat{\mathcal Y}$ from ground truth ${\mathcal Y}$ are available. The overall training objective is to find an encoding $ f_{V}(\cdot)$ and a $\bs{fc}$ layer for final predictions. Typical partial label learning has the form:
\begin{equation}\label{eq:formulation}
  \min_{\Theta}\mathbb E_{(\mathbf x,\hat{\mathbf y}) \sim \mathcal X \times \hat{\mathcal Y} }\xi( \bs{fc}(f_{V}(\mathbf x; \Theta)),\hat{\mathbf y}),
\end{equation}
where $\Theta$ is the model parameters for $f$ and $\xi:(0,1)^C \times \hat{\mathcal Y}$ is the optimization criterion,~\eg, binary cross entropy.

\textbf{Language-guided Multi-label Learning.} Beyond the methods with only visual features, text information serves as priors in prevailing methods~\cite{interactiveCNN_CVPR,Kundu2020exploitweakly,Wu2015MLMG} and our methods take the additional text inputs $\br{t} \in \mc{T}$ as inputs. In our framework (\Cref{fig:pipeline}), we encode the class-specific label space $\mc{T}$ of only $C$ word embeddings.  With the text encoder $f_{T}(\cdot ;\theta_{t})$, language-guided multi-label learning modifies \Cref{eq:formulation} as:
\begin{equation}\label{eq:formulation2}
  \min_{\Theta}\mathbb E_{(\mathbf x,\hat{\mathbf y}) \sim \mathcal X \times \hat{\mathcal Y} }\xi(\br{p}(\mathbf x,\br{t};\Theta),\hat{\mathbf y}),
\end{equation}
where $\Theta = \{\theta_{v},\theta_{t},\theta_{g}\}$, and $\br{p}(\mathbf x,\br{t};\Theta)=\bs{fc}(\mc{G}(f_{T}(\br{t};\theta_t)\rightarrow f_{V}(\mathbf x; \theta_v));\theta_{g})$ denotes our LDSA prediction. $\mc{G}(\cdot \rightarrow \cdot)$ denotes the guidance module. Hence our proposed method has the following objectives: 1) constructing accurate visual features by adapting CLIP visual encoders $f_{V}$; 2) adapting and transferring text priors to guide the learning process with $f_{T}$ and $\mc{G}$.

\subsection{Implicit Contrastive Learning}\label{sec:vision}
\textbf{From Sparse to Dense.} The most common methods including CLIP~\cite{clip} for classification are to embed the extracted feature $f_{V}(\mathbf x) \in \mathbb{R} ^{WH\times D}$ with a pooling operation to get a sparse vector,~\ie, $\br{V}^{s}=\frac{1}{WH} \sum_{i=1}^{WH}f_{V}(\mathbf x) \in \mathbb{R} ^{1\times D}$. $W,H$, and $D$ denote the feature width, height, and dimensions respectively. However, this encoding-pooling paradigm simply omits the spatial representations of different objects, which limits further object relation discovery and multi-modal interactions. We thus replace the data-dependent attention pooling in CLIP with dense features $\br{V}^{d}$, resulting in $WH$ local vectors with the size of  $\mathbb{R} ^{1\times D}$.

\begin{figure}[t]
  \centering
  \includegraphics[width=\linewidth]{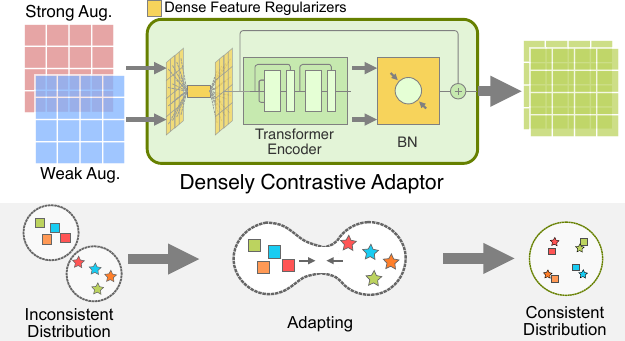}
  \caption{Illustrations of the densely contrastive adaptor. The adaptor consists of dense feature regularizers and transformer encoders, connected with residual connections. Initially, the strong and weak versions of the same image do not exhibit the expected correspondence in latent space, leading to inconsistent distribution.}
  \label{fig:cl-adapter}
\end{figure}

\textbf{Implicit Semantic Contrastive Relations.} After extracting the dense representations of different images, the intrinsic structures and semantic attributes are still under-explored. One common method to explore object-wise relationships is to exploit unsupervised Contrastive Learning~\cite{simclr} with strong or weak augmentations. As multi-label learning with partial annotations is a weakly supervised problem, here we directly use the semantic categories instead of the binary classification in standard InfoNCE constraints~\cite{oord2018infonce}. As in \Cref{fig:pipeline} and \Cref{fig:cl-adapter}, given a batch of training samples $\br{B}=\{\br{x}_{i}\}_{i=1}^{N}$, the standard optimization derived from \Cref{eq:formulation2} is $\mc{L}_{\bs{BCE}} = \frac{1}{N} \sum_{i=1}^{N} \xi(\br{p}(\mathbf x_i,\br{t}; \Theta),\hat{\mathbf y_{i}})$, where $\br{p}$ is the brief notation of the LDSA prediction. While our method conducts a strong $\mathrm{A}_s(\cdot)$ and weak augmentation $\mathrm{A}_w(\cdot)$ for a joint batch $\br{B}^{cl}=\{\mathrm{A}_s(\br{x}_{i}),\mathrm{A}_w(\br{x}_{i})\}_{i=1}^{N}$.
During each optimization, we conduct the implicit contrastive optimization of dual-stream features:
\begin{equation}\label{eq:contrastive}
\mc{L}_{\bs{IC}} = \frac{1}{2N} \sum_{i=1}^{N} (\xi(\br{p}(\mathrm{A}_s(\br{x}_{i}),\br{t}; \Theta),\hat{\mathbf y_{i}})+\xi(\br{p}(\mathrm{A}_w(\br{x}_{i}),\br{t}; \Theta),\hat{\mathbf y_{i}})).
\end{equation}
This learning stream explores the contrastive visual consistencies with the averaged gradients during optimization, which \textit{stands at a different view compared with common data augmentation operations}. We will discuss the relations by using InfoNCE and positive-aware L1 constraints in \Cref{sec:loss}. Moreover, this contrastive input works collaboratively with the proposed adaptors in the next paragraph.

\textbf{Densely Contrastive Adaptor.} Naive finetuning on CLIP-based visual backbones would face severe feature collapse and thus lead to catastrophic overfitting on limited-seen training labels. A naively fine-tuned CLIP only achieves $34.0$ mAP on the MS-COCO dataset. To overcome this problem, we freeze the pretrained visual encoder $f_{V}$ for downstream finetuning. Then a non-invasive densely contrastive adaptor is plugged after the encoder $f_{V}$. With the dense semantic features $\br{V}^{d}$, our proposed adaptor consists of two essential modules,~\ie, dense feature regularizers (\ie, dense encoder and dense batch norm layer) and transformer encoders in \Cref{fig:cl-adapter}.

To encode the dense semantic feature in a lightweight manner, we develop a weight-sharing encoder with the learnable weight $\br{W}_{\{s,e \}}$ and bias $\br{b}_{\{ s,e \}}$. For each dense feature $\br{V}^{d} \in \mathbb{R}^{|\br{B}^{cl}| \times WH \times C}$ of the $i$th image in joint batch $\br{B}^{cl}$, we have:
\begin{equation}
\br{G}^{d}_{ij}=\br{V}^{d}_{ij}+\mathbf W_{e}^\top\bs{ReLU}(\mathbf W_{s}^\top \br{V}^{d}_{ij}+\mathbf b_{s})+\mathbf b_{e}, 
\end{equation}
where $\mathbf W_s \in \mathbb R^{D \times D_b}, \mathbf W_e \in \mathbb R^{D_b \times D}, D_b<D$, which follows a Squeeze-and-Excitation encoding fashion only on the channel dimension. After that, a standard two-layer Transformer is adopted as a feature encoder:
\begin{equation}\begin{split}
\br{G}^{a}_{ijk}&=[\br{G}^{d}_{ijk}]+\bs{Att}^2(\mathbf W_{v}^\top[ \br{G}^{d}_{ijk}+\Delta(j,k)], \\
    &\qquad\quad\mathbf W_{q}^\top[ \br{G}^{d}_{ijk}+\Delta(j,k)])\cdot\mathbf W_{k}^\top[ \br{G}^{d}_{ijk}+\Delta(j,k)],
    \end{split}\end{equation}
where $\br{W}_{\{q,k,v\} }$ denotes the learnable weights for multi-head attention $\bs{Att}(\cdot,\cdot)$ and $\Delta(\cdot;\theta_\Delta): \mathbb N \times \mathbb N \mapsto \mathbb R^d$ is the learnable positional encoding.

As aforementioned, our motif is to optimize the joint batch $\br{B}^{cl}$ with dual augmentation views simultaneously. Thus we force the network to learn averaged statistics of dual views and improve the conventional batch norm from two aspects,~\ie, 1) dense normalization for spatial resolution, and 2) using joint statistics for each step optimization. Hence the mean $\mu$ and $\sigma$ of BN layers are computed as:
\begin{align}
  \mu_B&=\frac{1}{|\br{B}^{cl}| \times WH} \sum_{b=1}^{|\br{B}^{cl}|} \sum_{i=1}^{WH}  \br{G}^{a}_{b,i}, \\
  \sigma^2_B&=\frac{1}{|\br{B}^{cl}| \times WH} \sum_{b=1}^{|\br{B}^{cl}|} \sum_{i=1}^{WH}  (\br{G}^{a}_{b,i}-\mu_B)^2.
\end{align}
Then we follow the common normalization operation~\cite{Ioffe2015batchNorm} (affine values $\gamma,\beta$) with a residual input for final vision outputs $\br{H}$:
\begin{equation}
\br{H}_{b,i}=\br{G}^{a}_{b,i}+\gamma \frac{\br{G}^{a}_{b,i}-\mu_B}{\sqrt{\sigma_B+\epsilon}}+\beta.
\end{equation}
In this manner, the network tends to find a joint compact semantic space for weak and strong views of input images. Due to the patch-level semantic consistency and exchangeability of the attention operations, it implicitly constructs positive pairs between patches of two augmentations of one image, and negative pairs between patches of different images in a batch.

\begin{figure*}[t]
  \centering
  \includegraphics[width=0.9\linewidth]{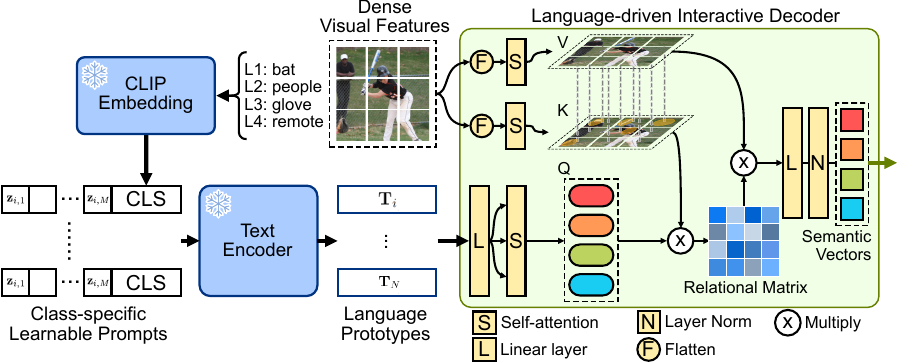}
  \caption{Illustration of language-guided learning. Our proposed method first uses class-specific learnable prompts to guide the alignment of visual and text domains. Then we construct a cross-modal interaction to guide the representation of dense visual features by exploiting the transformer architecture. We finally construct the compact semantic prediction instead of the widely used $\bs{fc}$ layers. }\label{fig:language-guide}
\end{figure*}

\subsection{Language-driven Interactive Decoder}\label{sec:text}
In \Cref{sec:vision}, we conduct a fast adaptation of vision modalities for downstream multi-label representations.
However, this individual tuning process would break the alignment of pretraining multimodal knowledge in CLIP (\Cref{fig:illstration}).
Hence we introduce a fast prompt learning strategy for modality alignment and then use the aligned semantic relationships to guide the visual representations.

\subsubsection{Class-specific Context Prompts}
Tuning downstream tasks with all network parameters in text encoders would also lead to feature collapse.
Here we resort to the wide-used prompt learning following~\cite{zhou2022coop,Rao2022denseclip,Zhou2022conditionalprompt}, which freezes the backbone encoder with only a learnable text prompt $\br{t}$.
Hence for each class, we construct $\br{t}_i= [\br{z}_{i,1}\ldots \br{z}_{i,M}, \bs{cls} ]$ with the length of $M$, where $\bs{cls}$ denotes the class text label,~\eg, $[\bs{person}]$.
We thus build text-based prototypes of $C$ classes towards the downstream dataset with fixed learning weights $\theta_{\mathrm{fix}}$: $\br{T}_{i} = f_{T}(\br{t}_i;\theta_{\mathrm{fix}})$.

\subsubsection{Cross-modal Interactive Decoding}
Predominant works~\cite{interactiveCNN_CVPR,Kundu2020exploitweakly,Wu2015MLMG} adopt the predetermined text-wise similarities to encourage the co-occurrence of fixed pairs,~\eg, simply enhancing the prediction of \textit{bat} classes if there is a \textit{person}.
Benefiting from the strictly aligned language and vision modalities, here we propose to construct the image-text cross relations with Transformer style decoders.
The language prototypes $\br{T}\in \mathbb{R}^{C\times D}$ of $C$ classes serve as query proxies to build the relation matrix $\br{A}$:
\begin{equation} \label{eq:att}
    \br{A}_{c,j} =(\br{T}_{c,i}+\Delta(c,i))\mathbf W^{d}_{q} \mathbf {W}^{d}_{k} (\br{H}_{i,j}+\Delta(i,j)) \in \mathbb{R}^{C\times WH},
\end{equation}
where $\mathbf W^{d}_{ \{ q,k \} }$ are learnable dimensional transformation weights.
This relational matrix $(\mathbb{R}^{C} \times \mathbb{R}^{WH})$ denotes the semantic correlations of each class and spatial region patches.
With this cross-modality relationship, we then use the multi-head attention $\bs{Att}(\cdot,\cdot)$ to obtain the final output:
\begin{equation} \label{eq:final} 
\br{O}_{c,i}=\bs{Att}(\bs{Norm}(\br{A}_{c,j}), (\br{H}^{\top}_{i,j}+\Delta(i,j)) \mathbf{W^{d}_{v}}^{\top}).
\end{equation}

We plot the details in \Cref{fig:language-guide}. We freeze the backbone encoder $f_T$ and its token embeddings \texttt{cls}, and resort to prompt learning where we construct class-specific learnable prompts $\mathbf t_i$ for each label to build language prototypes $\mathbf T_i \in \mathbb R^D$ towards the downstream dataset:
\begin{equation}
  \mathbf T_i=f_T([\mathbf z_{i,1},\dots, \mathbf z_{i,M}, \mathtt{cls}];\theta_{\mathrm{fix}}),
\end{equation}
where $D$ is the size of feature channels. The language prototypes $\{\mathbf T_i\}_{i=1}^C$ are then used in the language-driven interactive decoder, serving as query proxies. The key and value inputs of the decoder come from dense visual features, which indicate the images and their spatial contexts. We next construct the image-text cross relational matrix $\mathbf A$ and obtain the semantic vectors $\mathbf O$ for prediction.

\subsubsection{Semantic-aware Compact Prediction}
After transferring the cross-modality relationship to the dense visual embedding, the final output is constructed of semantic tensor groups,~\ie,$\br{O}\in \mathbb{R}^{C\times D}$.
It can be deemed as $C$ semantic vectors with the dimension of $D$, and each vector has an explicit correlation with specific categories.
Thus we construct $C$ independent networks $\tau:\mathbb{R}^{D}\rightarrow \mathbb{R}^{1}$ to predict the existence of each category:
\begin{equation} \label{eq:pred} 
\br{p} = \bs{Concat}(\tau_{1}(\br{O}_{1,\cdot})\ldots \tau_{C}(\br{O}_{C,\cdot})).
\end{equation}
In this way, we construct a compact semantic mapping with $C\times D\times 1$ parameters to optimize, while the conventional $\bs{fc}$ layer requires $C D\times C$ for optimization.
This compact training with only $\frac{1}{C}$ parameters notably regularizes the feature representation in a generalized space.

\subsection{Implicit Semantic Contrastive Learning}\label{sec:loss}
As mentioned above, conventional contrastive learning usually relies on the binary regularization of negative and positive pairs, here we discuss the relations and derivation with commonly used learning objectives.

\textbf{Derivations from Positive L1 and InfoNCE.} We denote the predictions of the weak and strong views of the same image as $\br{p}^{w}_i=\br{p}(\mathrm{A}_w(\br{x}_i),\br{t};\Theta)$ and $\br{p}^{s}_i=\br{p}(\mathrm{A}_s(\br{x}_i),\br{t};\Theta)$, respectively. The positive L1 aims to regularize the similarity of these paired predictions. Thus we have the positive L1: $ \mc{L}_{\bs{P-1}}(\{\br{p}^w_i,\br{p}^s_i\};\Theta)=\frac 1 N \sum_{i=1}^N\lVert\br{p}^{w}_i - \br{p}^{s}_i\rVert_1.$ We denote this regularization as $\mc{L}_{\bs{P-1}}=\frac 1N \sum_{i=1}^N(-a_+)$ for simplicity, which only constrains the similarity of positive relationships. Based on this foundation, we construct InfoNCE~\cite{oord2018infonce} as $\mc{L}_{\bs{Info}}=\frac{1}{M}\sum_{i=1}^N\left(- b_+ +  \log\sum_{j=1}^N b_- \right)$, which incorporates an additional negative term:  
\begin{equation}\label{eq:infonce}
\mc{L}_{\bs{Info}}=\frac{1}{2N}\sum_{i=1}^N\left(- \rho(\br{p}^w_i, \br{p}^s_i)  +  \log\sum_{j}^{j\ne i} \text{e}^{\rho(\br{p}^w_i,\br{p}^s_j)} \right),
\end{equation}
where $\rho(\cdot)$ denotes the cosine similarity.

\textbf{Implicit Semantic Contrastive Learning.} Although the \Cref{eq:infonce} successfully forms the positive and negative clusters, the semantic representations and relationships are neglected. Thus we incorporate the Asymmetric Loss \cite{Ridnik2021asymmetricloss} into our framework with semantic contrastive regularization:
\begin{equation}
\begin{split}
&\mc{L}_{\bs{ISC}}=-\frac{1}{2N}\sum_{i=1}^N\sum_{j=1}^C\left(\mathds 1_{\hat{\br{y}}_{ij}=1}\mc{L}_{+} + \mathds 1_{\hat{\br{y}}_{ij}=-1}\mc{L}_- \right) \\
    &=\frac{1}{2N}\sum_{i=1}^N\left(-\sum_{\substack{1 \leq j \leq C \\ \mathrm{s.t.}~\hat{\br{y}}_{ij}=1}} c_++\sum_{\substack{1 \leq j \leq C \\ \mathrm{s.t.}~\hat{\br{y}}_{ij}=-1}} c_-\right).
\end{split}
\end{equation}
Here we extend the prediction $\br{p}_{ij}$ in the image dimension to the semantic category dimension indexed by $j$, and optimize it against the partial label $\hat{\br{y}}_{ij}$. By ignoring the unknown annotations ($\hat{\br{y}}_{ij}=0$), we attach different importance here:
\begin{equation}
\begin{cases}
    c_+&=\log \sigma(\br{p}^{w}_{ij})\sigma(\br{p}^{s}_{ij}), \\
    c_-&=\log\sigma(-\br{p}^{w}_{ij})^{-1}\sigma(-\br{p}^{s}_{ij})^{-1},
\end{cases}
\end{equation}
where $\sigma(\cdot)$ denotes the Sigmoid function.

We argue that during this implicit learning process, the supervised multi-label task and semantic contrastive relationship are jointly optimized within this unified objective.
The final training objective function for our proposed LDSA is:
\begin{equation}\label{eq:total_loss}
    \mathcal{L}_{\mathrm{total}} = \mathcal{L}_{\bs{ISC}},
\end{equation}
which simultaneously minimizes the classification error and regularizes the semantic space through the dual-view contrastive mechanism.

\textbf{Optimization Algorithms.} With the proposed implicit learning, the training process of the proposed Language-driven Dense Semantic Adaptor is elaborated in \Cref{alg:two}, where we describe: 
\begin{enumerate}
    \item The initialization of networks, including fixed encoder $f_{\{V,T\}}$, prompts $\mathbf t_i$, the densely contrastive adaptor, and the language-driven interactive decoder;
    \item The construction of a contrastive batch, which contains weak and strong augmented views of the same series of images; 
    \item The pipeline, including details of prompt learning, visual/textual encoding, language-driven interactive decoding, and predicting.
\end{enumerate}

\begin{figure}[t]
  \centering
  \includegraphics[width=\linewidth]{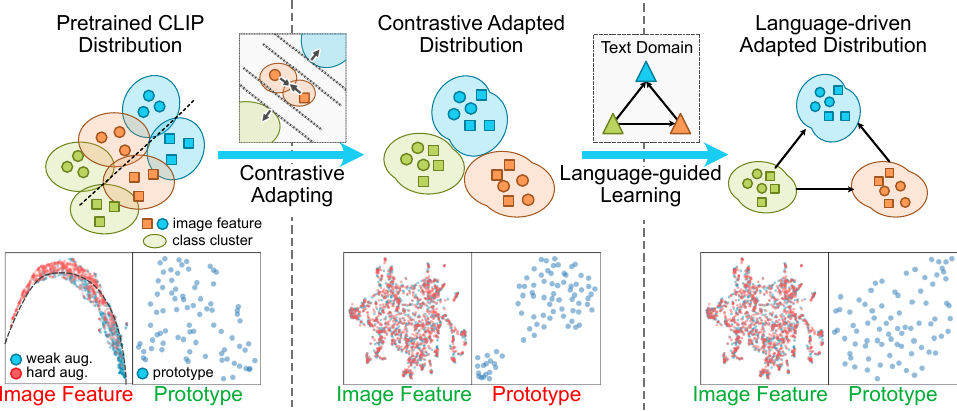}
  \caption{The illustration of feature space alignment with proposed LDSA on the MS-COCO dataset, where the distributions of the real image feature and prompt prototypes are visualized for each step. 1) With pretrained CLIP, prototypes are distributed {\color{green}wisely} while contrastive weak and hard views of images are {\color{red}incorrectly} distributed in more than one manifold. 2) After adapting, image features are {\color{green}correctly} clustered, however, the prototypes are {\color{red}corrupted}. 3) With language-guided learning, LDSA learns the {\color{green}optimal} representations, capturing the correspondence of images, as well as relationships between labels.}
  \label{fig:illstration}
\end{figure}

\textbf{Discussion of Feature Alignment.} With the proposed LDSA, multimodal features of CLIP are aligned into the downstream data by contrastive adaptation and language-driven learning. In \Cref{fig:illstration}, we illustrate the distribution of image features, derive the distribution of image features, and label prompt prototypes from the MS-COCO dataset. The pretrained CLIP does not recognize the correspondence between strong and weak views of images, however, the distribution of prototypes is uniform, suggesting that CLIP retains rich knowledge of labels. With the Densely Contrastive Adaptor, we successfully adapt the visual representations. However, individual adaptation of visual features leads to misalignment. By introducing both adaptation and language-driven learning, LDSA finally achieves adapting CLIP to downstream tasks while retaining its knowledge for adapting relationships, where both image features and semantic prototypes are well distributed in space.

\begin{algorithm}[t]
  \caption{Language-driven Dense Semantic Adaptor Training Algorithms}\label{alg:two}
  \KwData{Dataset $\mathcal D=(\mathcal X,\hat{\mathcal Y}).$}
  \KwResult{prompts $\{t_c\}_{c=i}^C$; adaptor and decoder.}
  Initialize visual encoder $f_V$ and text encoder $f_T$ from CLIP \\
  Random Init. prompts $\{\mathbf t_c\}_{c=1}^C$ \\
  Random Init. densely contrastive adaptor \\
  Random Init. language-driven interactive decoder \\
  \For{$\mathbf B \sim \mathcal D$}{
    {\color{blue}/* Augmenting */} \\
    $\mathbf B^{cl}=\{\mathrm{Aug_w}(\mathbf B),\mathrm{Aug_s}(\mathbf B)\}$ \\
    {\color{blue}/* Visual encoding */} \\
    \For{$(\mathbf x_i,\hat{\mathbf y}_i) \in \mathbf B^{cl}$}{
      $\mathbf V^d_i=f_V(\mathbf x_i;\theta_{\mathrm{fix}})$ \\
      $\mathbf G^d_{ijk}=\mathbf V^d_{ijk}+W^\top_e \mathtt{ReLU}(\mathbf W_s^\top\mathbf V^d_{ijk}+\mathbf b_s)+\mathbf b_e$ \\
      $\mathbf G^d_{ijk}=\mathbf G^d_{ijk}+\mathtt{Att}^2(\mathbf W^\top_v (\mathbf G_{ijk}^d+\Delta(j,k)), \mathbf W_q^\top (\mathbf G_{ijk}^d+\Delta(j,k)))\cdot \mathbf W_k^\top(\mathbf G_{ijk}^d+\Delta(j,k))$ \\
    }
    $\mu_B=\frac{1}{|\mathbf B^cl| \times WH}\sum_{i=1}^{|\mathbf B^{cl}|}\sum_{j=1,k=1}^{W,H}G_{ijk}^a$ \\
    $\sigma_B=\frac{1}{|\mathbf B^cl| \times WH}\sum_{i=1}^{|\mathbf B^{cl}|}\sum_{j=1,k=1}^{W,H}(G_{ijk}^a-\mu_B)$ \\
    $\mathbf H_{ijk}=\mathbf G^a_{ijk}+\gamma\frac{\mathbf G^a_{ijk}-\mu_B}{\sqrt{\sigma_B+\epsilon}}+\beta$ \\
    {\color{blue}/* Textual encoding */} \\
    $\mathbf T_c=f_T(\mathbf t_c;\theta_{\mathrm{fix}})$ \\
    {\color{blue}/* Interactive decoding */ }\\
    $\mathbf O_{ic}=\mathtt{InteractiveDecoder(\mathbf H_{ijk},\mathbf T_c)}$ \\
    {\color{blue}/* Compact predicting */ } \\
    $\mathbf p_i=\mathtt{Concat}(\tau_1(\mathbf O_{i1}),\dots,\tau_1(\mathbf O_{iC}))$ \\
    {\color{blue}/* Optimization */} \\
    Calculate the joint loss: $\mathcal{L}_{total} = \mathcal{L}_{\bs{ISC}}$ \;
    Update parameters: $\Theta \leftarrow \Theta - \eta \nabla_{\Theta} \mathcal{L}_{total}$ \;
  }
\end{algorithm}

\section{Experiments} \label{sect:exp}

\subsection{Datasets and Evaluation Metrics}

\textbf{Datasets.} Following previous studies \cite{sst_aaai2022,Pu2022SARB}, we conduct experiments on three representative benchmarks, \ie, PASCAL VOC 2007 \cite{pascal-voc-2007}, Microsoft COCO 2014 \cite{coco}, and Visual Genome \cite{krishna2017visual}. These fully labeled datasets are widely used in multi-label image classification research. PASCAL VOC 2007 contains $20$ categories, where each image in the dataset has $1.4$ labels on average. MS-COCO is the most widely used and challenging benchmark in multi-label classification tasks, which contains $80$ categories. Each image in the MS-COCO dataset contains 2.9 labels on average. Visual Genome contains 108K images and various categories. We follow \cite{Pu2022SARB} to filter 200 most frequent labels to obtain VG-200 dataset and partition it into training and test sets following the conventional 7:3 ratio. Since all datasets are fully annotated, following \cite{sst_aaai2022,Pu2022SARB}, we randomly drop labels in training sets to construct partial label datasets. The details are listed in \Cref{sec:data-preprocessing}.

\textbf{Evaluation Metrics.} For fair comparisons, we follow previous studies \cite{sst_aaai2022,Pu2022SARB} to adopt mean average precision (mAP) as our main evaluation metrics. The results are reported on official validation sets with different amounts of labels known in training sets, of which proportions are set as $10\%,20\%,\dots,90\%$ respectively.

\subsection{Implementation Details}

\textbf{Data Preprocessing.}\label{sec:data-preprocessing} Following the partial label setting \cite{sst_aaai2022,Pu2022SARB}, we first randomly drop existing labels of the training set, of which the proportions are respectively $10\%,20\%,\dots,90\%$, resulting in partially labeled datasets with known labels of different proportions generated from fully labeled PASCAL VOC, MS-COCO and VG-200 datasets. Such operation is performed once and all approaches are trained with the same retained labels. For performance evaluation, we keep all annotations of the official validation sets, and the results of ours and other approaches are reported on them.

\textbf{Training Details.}
We adopt CLIP \cite{clip} with ResNet-101 pretrained for fair comparisons. Following previous works \cite{sst_aaai2022,Pu2022SARB,mlgcn_CVPR2019,Durand_2019_CVPR}, images are resized into $448 \times 448$ with data augmentation. We implement learnable positional encoding for densely contrastive adaptor following \cite{detr}. We adopt the AdamW optimizer and train the model for $6$ epochs. The learning rates for the adaptor and decoder are set as $5 \times 10^{-5}$. We adopt a learning rate scheduler that decreases the learning rate to $5 \times 10^{-6}$ at the third epoch. The CLIP backbone is fixed during training. The batch size is set as $64$. The dimension of visual and textual features is set as $512$. The parameter $D_b$ is set as $128$. The transformers consist of $2$ blocks with attention heads set as $2$, and the hidden dimension is set as $2048$. In addition, we follow PU-MLC \cite{yuan2024positivelabelneedmultilabel} and use the Exponential Moving Average (EMA) with a momentum of $0.999$ for the main experiment and $1.0$ for the ablation studies to rigorously evaluate the contribution of our methods. All experiments are conducted on a single NVIDIA 3090 GPU.

The weak and strong augmentations employed in our method are implemented as follows:
i) Weak augmentation consists of two basic operations: random horizontal flipping and random cropping (with scale ratios of $30\%$, $50\%$, and $60\%$ relative to the original image).
ii) Strong augmentation consists of the weak augmentation plus the RandAugment \cite{cubuk2020randaugment} strategy.

\subsection{Comparison with State-of-the-art}

\begin{table*}[t]
  \caption{Experimental comparison using mAP (\%) with the state-of-the-art methods on the PASCAL VOC, MS-COCO, and VG-200 benchmarks.}
  \label{tab:mAP-comparison}
  \centering
  \begin{tabular}{c|c||ccccccccc|c}
  \toprule
  \textbf{Benchmarks} & \textbf{Methods} & $10\%$ & $20\%$ & $30\%$ & $40\%$ & $50\%$ & $60\%$ & $70\%$ & $80\%$ & $90\%$  & \textbf{Avg.}\\
  \hline
  \multirow{13}{*}{MS-COCO \cite{coco} } & SSGRL \cite{Chen2019SSGRL} & $62.5$ & $70.5$ & $73.2$ & $74.5$ & $76.3$ & $76.5$ & $77.1$ & $77.9$ & $78.4$  & $74.1$ \\
   & GCN-ML \cite{Chen2019GCNML}  & $63.8$ & $70.9$ & $72.8$ & $74.0$ & $76.7$ & $77.1$ & $77.3$ & $78.3$ & $78.6$  & $74.4$\\
   & KGGR \cite{Chen2022KGGR}  & $66.6$ & $71.4$ & $73.8$ & $76.7$ & $77.5$ & $77.9$ & $78.4$ & $78.7$ & $79.1$ & $75.6$ \\
   & Curriculum Learning \cite{Durand_2019_CVPR}  & $26.7$ & $31.8$ & $51.5$ & $65.4$ & $70.0$ & $71.9$ & $74.0$ & $77.4$ & $78.0$  & $60.7$\\
   & Partial BCE \cite{Durand_2019_CVPR}  & $61.6$ & $70.5$ & $74.1$ & $76.3$ & $77.2$ & $77.7$ & $78.2$ & $78.4$ & $78.5$  & $74.7$\\
   & SST \cite{sst_aaai2022}  & $68.1$ & $73.5$ & $75.9$ & $77.3$ & $78.1$ & $78.9$ & $79.2$ & $79.6$ & $79.9$  & $76.7$\\
   & SST*  & $69.1$ & $78.5$ & $79.3$ & $79.9$ & $80.1$ & $80.5$ & $81.1$ & $80.7$ & $80.7$  & $78.9$\\
   & SARB \cite{Pu2022SARB}  & $71.2$ & $75.0$ & $77.1$ & $78.3$ & $78.9$ & $79.6$ & $79.8$ & $80.5$ & $80.5$  & $77.9$\\
   & SARB*  & $75.5$ & $78.5$ & $79.0$ & $79.5$ & $80.4$ & $80.2$ & $80.8$ & $80.6$ & $80.8$  & $79.4$\\
   & DualCoOp* \cite{sun2022dualcoop}  & $78.7$ & $80.9$ & $81.7$ & $82.0$ & $82.5$ & $82.7$ & $82.8$ & $83.0$ & $83.1$  & $81.9$\\
   & DualCoOp++* \cite{hu2023dualcoopfasteffectiveadaptation}  & $\underline{81.4}$ & $\underline{83.1}$ & $\underline{83.7}$ & $\underline{84.2}$ & $\underline{84.4}$ & $\underline{84.5}$ & $\underline{84.8}$ & $\underline{85.0}$ & $\underline{85.1}$  & $\underline{84.0}$\\
   & DualCoOp*+EMA \cite{sun2022dualcoop}  & $78.3$ & $78.3$ & $77.2$ & $78.7$ & $79.0$ & $79.3$ & $78.7$ & $79.8$ & $77.6$  & $78.5$\\
   & Ours & $\mathbf{83.9}$ & $\mathbf{85.4}$ & $\mathbf{86.3}$ & $\mathbf{86.6}$ & $\mathbf{87.1}$ & $\mathbf{87.6}$ & $\mathbf{87.8}$ & $\mathbf{88.0}$ & $\mathbf{88.2}$ & $\mathbf{86.8}$\\

   \hline
  \multirow{11}{*}{PASCAL VOC \cite{pascal-voc-2007} } & SSGRL \cite{Chen2019SSGRL} & $77.7$ & $87.6$ & $89.9$ & $90.7$ & $91.4$ & $91.8$ & $91.9$ & $92.2$ & $92.2$  & $89.5$\\
  & GCN-ML \cite{Chen2019GCNML}  & $74.5$ & $87.4$ & $89.7$ & $90.7$ & $91.0$ & $91.3$ & $91.5$ & $91.8$ & $92.0$  & $88.9$\\
  & KGGR \cite{Chen2022KGGR}  & $81.3$ & $88.1$ & $89.9$ & $90.4$ & $91.2$ & $91.3$ & $91.5$ & $91.6$ & $91.8$  & $89.7$\\
  & Curriculum Learning \cite{Durand_2019_CVPR}  & $44.7$ & $76.8$ & $88.6$ & $90.2$ & $90.7$ & $91.1$ & $91.6$ & $91.7$ & $91.9$  & $84.1$\\
  & Partial BCE \cite{Durand_2019_CVPR}  & $80.7$ & $88.4$ & $89.9$ & $90.7$ & $91.2$ & $91.8$ & $92.3$ & $92.4$ & $92.5$  & $90.0$\\
  & SST \cite{sst_aaai2022}  & $81.5$ & $89.0$ & $90.3$ & $91.0$ & $91.6$ & $92.0$ & $92.5$ & $92.6$ & $92.7$  & $90.4$\\
  & SARB \cite{Pu2022SARB} & $83.5$ & $88.6$ & $90.7$ & $91.4$ & $91.9$ & $92.2$ & $92.6$ & $92.8$ & $92.9$  & $90.7$\\
  & DualCoOp* \cite{sun2022dualcoop} & $90.3$ & $92.2$ & $92.8$ & $93.3$ & $93.6$ & $93.9$ & $94.0$ & $94.1$ & $94.2$  & $93.2$\\
  & DualCoOp++* \cite{hu2023dualcoopfasteffectiveadaptation} & $\mathbf{92.7}$ & $\underline{93.4}$ & $\underline{93.8}$ & $\underline{94.0}$ & $\underline{94.3}$ & $\underline{94.4}$ & $\underline{94.4}$ & $\underline{94.7}$ & $\underline{94.9}$  & $\underline{94.1}$\\
  & DualCoOp*+EMA \cite{sun2022dualcoop} & $90.8$ & $92.0$ & $92.9$ & $93.6$ & $93.0$ & $93.0$ & $92.6$ & $91.0$ & $91.0$  & $92.2$\\
  & Ours & $\underline{91.7}$ & $\mathbf{93.6}$ & $\mathbf{94.5}$ & $\mathbf{94.6}$ & $\mathbf{95.2}$ & $\mathbf{95.1}$ & $\mathbf{95.3}$ & $\mathbf{95.4}$ & $\mathbf{95.6}$  & $\mathbf{94.6}$\\

  \hline
    \multirow{9}{*}{VG-200} & SSGRL \cite{Chen2019SSGRL} & 34.6 & 37.3 & 39.2 & 40.1 & 40.4 & 41.0 & 41.3 & 41.6 & 42.1 & 39.7 \\
    & GCN-ML \cite{Chen2019GCNML} & 32.0 & 37.8 & 38.8 & 39.1 & 39.6 & 40.0 & 41.9 & 42.3 & 42.5 & 39.3 \\
    & KGGR \cite{Chen2022KGGR} & 36.0 & 40.0 & 41.2 & 41.5 & 42.0 & 42.5 & 43.3 & 43.6 & 43.8 & 41.5 \\
    & Curriculum labeling \cite{Durand_2019_CVPR} & 12.1 & 19.1 & 25.1 & 26.7 & 30.0 & 31.7 & 35.3 & 36.8 & 38.5 & 28.4 \\
    & Partial BCE \cite{Durand_2019_CVPR} & 27.4 & 38.1 & 40.2 & 40.9 & 41.5 & 42.1 & 42.4 & 42.7 & 42.7 & 39.8 \\
    & SST \cite{sst_aaai2022} & 38.8 & 39.4 & 41.1 & 41.8 & 42.7 & 42.9 & 43.0 & 43.2 & 43.5 & 41.8 \\
    & HST \cite{chen2024heterogeneous} & 40.6 & 41.6 & 43.3 & 44.6 & 45.2 & 45.8 & 46.8 & 47.2 & 47.8 & 44.8 \\
    & SARB \cite{Pu2022SARB} & $\underline{40.6}$ & $\underline{43.5}$ & $\underline{44.5}$ & $\underline{45.3}$ & $\underline{46.0}$ & $\underline{47.1}$ & $\underline{47.2}$ & $\underline{47.8}$ & $\underline{48.1}$ & $\underline{45.6}$ \\
    & Ours & $\mathbf{51.4}$ & $\mathbf{53.5}$ & $\mathbf{54.3}$ & $\mathbf{54.9}$ & $\mathbf{55.3}$ & $\mathbf{55.8}$ & $\mathbf{56.0}$ & $\mathbf{56.2}$ & $\mathbf{56.4}$  & $\mathbf{54.9}$\\
  \bottomrule
  \multicolumn{12}{l}{*: CLIP backbone.}
  \end{tabular}
\end{table*}

We respectively compare our approach on MS-COCO \cite{coco}, PASCAL VOC \cite{pascal-voc-2007} and VG-200 \cite{krishna2017visual} datasets with 10 state-of-the-art methods, including SSGRL \cite{Chen2019SSGRL}, GCN-ML \cite{Chen2019GCNML}, KGGR \cite{Chen2022KGGR}, Curriculum Learning \cite{Durand_2019_CVPR}, Partial BCE \cite{Durand_2019_CVPR}, SST \cite{sst_aaai2022}, SARB \cite{Pu2022SARB}, HST \cite{chen2024heterogeneous}, DualCoOp \cite{zhou2022coop}, and DualCoOp++ \cite{hu2023dualcoopfasteffectiveadaptation}. We additionally replace the backbone of SST and SARB with CLIP for fairness. \Cref{tab:mAP-comparison} lists the comparison of mAP between ours and all the other methods trained with $10\%$ to $90\%$ proportions of known labels on training sets. Our proposed LDSA surpasses the state-of-the-art models by a large margin on all proportions. The performance on MS-COCO and VG-200 is improved by $2.8\%$ and $9.3\%$ on average compared with the state-of-the-art DualCoOp++ and SARB. It achieves comparable performance on the VOC dataset, showing advantages at more various label proportions.

Besides the most widely used mAP, we report average overall precision (OP), overall recall (OR), overall F1-score (OF1), per-class precision (CP), per-class recall (CR), and per-class F1-score of ours and other methods in \Cref{tab:evaluation-metrics}. The results show that our proposed method keeps a balance between precisions and recalls, greatly improving F1 scores on MS-COCO and all metrics on PASCAL VOC and VG-200. DualCoOp \cite{sun2022dualcoop} gains better OR and CR metrics on MS-COCO, benefiting from the general semantic knowledge of CLIP \cite{clip}. However, its precisions are lower, which we believe can be attributed to the incomplete transfer of the knowledge that is for the classification of the downstream datasets.

\begin{table}[t]
  \centering
  \caption{Ablation study for different modules on MS-COCO dataset. \\ DF: Resolution of Dense Feature. LG: Language Guidance. }
  \label{tab:component-ablation}
    \setlength{\tabcolsep}{3.0mm}
  \renewcommand{\arraystretch}{1.0}
  \begin{tabular}{cccc|cc}
  \toprule
  \textbf{DF} & \textbf{Adaptor} & \textbf{LG} & \textbf{Prompt} & \textbf{mAP} & $\Delta$\\
  \midrule
  \multicolumn{4}{l|}{Fine-tuned CLIP} & 34.0 & $\textcolor{black}{(-33.3)}$\\
  \hline
  \multicolumn{4}{l|}{+BYOL} & 71.2 & $(+3.9)$\\
  \multicolumn{4}{l|}{+MoCo} & 75.0 & $(+7.7)$\\
  \multicolumn{4}{l|}{+DetCo} & 59.2 & $(-8.1)$\\
  \multicolumn{4}{l|}{+UP-DETR} & 68.5 & $(+1.2)$\\
  \multicolumn{4}{l|}{+ORL} & 39.7 & $(-27.6)$\\
  \hline
  $\times$ & & & & $67.3$ & $(+0.0)$ \\
  $14$ & & & & $69.8$ & $(+2.5)$ \\
  $14$ & $\checkmark$ & & & $73.2$ & $(+5.9)$ \\
  $14$ & & $\checkmark$  &  & $77.0$& $(+9.7)$ \\
  $14$ & & $\checkmark$ & $\checkmark$  & $77.1$ & $(+9.8)$ \\
  $14$ & $\checkmark$ & $\checkmark$ & $\checkmark$ & $79.1$& $(+11.8)$ \\
  \hline
  $28$ & $\checkmark$ & $\checkmark$ & $\checkmark$ & $80.1$& $(+12.8)$ \\
  $7$ & $\checkmark$ & $\checkmark$ & $\checkmark$ & $76.7$& $(+9.4)$ \\
  $4$ & $\checkmark$ & $\checkmark$ & $\checkmark$ & $68.2$& $(+0.9)$ \\
  $2$ & $\checkmark$ & $\checkmark$ & $\checkmark$ & $49.9$& $(-17.4)$ \\

  \bottomrule
  \end{tabular}
\end{table}

\subsection{Performance Analysis}\label{sec:exp_per}

\textbf{Ablation Study for LDSA.} We conduct detailed ablations in \Cref{tab:component-ablation}, to meticulously analyze the impacts of various factors on $10\%$ partial setting with EMA disabled, including the components, dense feature resolutions, and object-level enhancements.

While a naively fine-tuned CLIP only achieves $34.0$ mAP on the MS-COCO dataset, the baseline model reaches the performance of $67.3$. However, after enabling dense representations coupled with visual adaptors, the performance climbs to $73.2$. This improvement underscores the pivotal role that both dense representations and visual adaptors play in refining the baseline model, indicating a synergy between these components and the core in understanding and processing visual content more effectively.

High-resolution dense features outperform their lower-resolution counterparts, demonstrating reliability in multi-label learning. While lower resolutions result in the loss of spatial information and fine-grained details, excessively high resolutions, despite preserving detail, introduce background noise and increase computational overhead. In practice, a relatively high resolution should be prioritized to strike a balance between performance and efficiency.

By further introducing the multi-modal language guidance, the performance shows a relative improvement of $10.8\%$ in the fourth row. The application of multi-modal language guidance demonstrates its effectiveness in bridging the semantic gap between different modalities, thereby enhancing the model's ability to leverage contextual cues from the language domain to inform its visual understanding. The incremental improvements observed at each step of the ablation verify the significance of our method.

To learn the effect of other object-level or contrastive learning methods, we further replace our LDSA with object-level methods DetCo \cite{xie2021detco}, ORL \cite{xie2021unsupervised}, UP-DETR \cite{dai2022updetr}, BYOL \cite{byol} and MoCo \cite{moco_CVPR2020} in \Cref{tab:component-ablation}. We adopt their official pre-trained model and fine-tune them following the same baseline. Although these methods leverage object detection, the discovery of object-level relationships on partial-label datasets is still challenging. Generic contrastive methods such as BYOL and MoCo fail to overcome the semantic inconsistency of multi-label images and remain below LDSA.

\begin{table}[!t]
  \caption{Ablation study for detailed designs of densely contrastive adaptor on MS-COCO dataset.}
  \label{tab:ablation-adapter}
\setlength{\tabcolsep}{3.3mm}
  \renewcommand{\arraystretch}{1.0}
  \centering
  \begin{tabular}{cc|cc}
  \toprule
  \textbf{Adaptor} & \textbf{Augumentations} & \textbf{mAP} & $\Delta$\\
  \midrule
  $\times$ & weak & $77.1$ & $\textcolor{black}{(+0.0)}$  \\
  $\times$ & strong  &  $74.1$ & $\textcolor{blue}{(-3.0)}$ \\
  $\times$ & weak + strong & $77.9$ & $\textcolor{red}{(+0.8)}$ \\
  \hline
  $\checkmark$ & weak &  $77.9$ & $\textcolor{red}{(+0.8)}$ \\
  $\checkmark$ & strong & $70.8$ & $\textcolor{blue}{(-6.3)}$ \\
  $\checkmark$ & weak + strong &  $\mathbf{79.1}$ & $\textcolor{red}{(+2.0)}$ \\
  \bottomrule
  \end{tabular}
\end{table}

\textbf{Ablation Study for Densely Contrastive Adaptor.} \label{sect:ablation-adaptor} We ablate the details of the adaptor in \Cref{tab:ablation-adapter} to find out how the adaptor influences the overall performance of our model. The \textbf{regularizers} denotes the dense weight-sharing encoder, with dense contrastive BNs.

We observe that augmentations and adaptors are mutually dependent rather than straightforwardly additive. The performance improvement observed only when both are enabled points to an implicit contrastive mechanism facilitated by the adaptor. Without the adaptor, strong augmentations act as noise; with it, they form informative contrastive pairs. This synergy empowers the model to effectively capture intricate features, indicating that the adaptor is essential for stabilizing and exploiting aggressive augmentation strategies, thereby enhancing the model's ability to capture and utilize complex patterns.

\begin{table}[t]
  \centering
  \caption{Detailed ablation study for dense feature regularizers.\\ DE: Dense encoder. BN: Dense batch norm.}\label{tab:ablate-regularizers}
    \setlength{\tabcolsep}{3.0mm}
  \renewcommand{\arraystretch}{1.0}
  \begin{tabular}{cc|cc||cc}
  \toprule
  \multicolumn{2}{c|}{\textbf{Regularizers}} & \multicolumn{2}{c||}{\textbf{Augmentations}} & \multirow{2}{*}{\textbf{mAP}} & \multirow{2}{*}{$\Delta$} \\
  \cline{1-4}
  \textbf{DE} & \multicolumn{1}{c|}{\textbf{BN}} & \textbf{Weak} & \textbf{Strong} & \\
  \midrule
   &  & $\checkmark$ &  & $77.1$ & $\textcolor{black}{(+0.0)}$\\
  \midrule
  $\checkmark$ &  & $\checkmark$ &  & $77.9$ & $\textcolor{red}{(+0.8)}$\\
  $\checkmark$ &  &  & $\checkmark$ & $76.3$ & $\textcolor{blue}{(-0.8)}$\\
  $\checkmark$ &  & $\checkmark$ & $\checkmark$ & $78.6$ & $\textcolor{red}{(+1.5)}$\\
   & $\checkmark$ & $\checkmark$ &  & $77.2$ & $\textcolor{red}{(+0.1)}$\\
   & $\checkmark$ &  & $\checkmark$ & $75.7$ & $\textcolor{blue}{(-1.4)}$\\
   & $\checkmark$ & $\checkmark$ & $\checkmark$ & $78.2$ & $\textcolor{red}{(+1.1)}$\\
  \midrule
   $\checkmark$ & $\checkmark$ & $\checkmark$ & $\checkmark$ & $\mathbf{79.1}$ & $\textcolor{red}{(+2.0)}$\\
  \bottomrule
  \end{tabular}
\end{table}

\textbf{Ablation Study for Dense Feature Regularizers.}\label{sec:ablation-regularizers} The densely contrastive adaptor mainly consists of transformer encoders and dense feature regularizers, of which the latter contains the dense encoder and dense batch norm layer.

To evaluate the specific impact of the dense feature regularizers, we futher conduct a detailed ablation study in \Cref{tab:ablate-regularizers}. Notably, the removal of any component leads to a performance degradation. This is especially evident when applying strong augmentations alone, where the absence of regularizers causes the model to struggle with high-variance noise. These findings suggest that the regularizers provide structural regularization over the feature space, effectively aligning diverse augmented data into a coherent representation. By enforcing this structural constraint, the regularizers enable the model to distill consistent patterns from data distributions.

\begin{table}[!t]
  \centering
  \caption{Ablation study for semantic-aware compact prediction. CP: Semantic-aware compact prediction. AP: Average prediction. FP: Fully-connected prediction.}\label{tab:ablation-prediction}
  \setlength{\tabcolsep}{1.5mm}
  \renewcommand{\arraystretch}{1.0}
  \begin{tabular}{c|ccccccc}
  \toprule
  \textbf{Methods} & \textbf{mAP} & \textbf{OP} & \textbf{OR} & \textbf{OF1} & \textbf{CP} & \textbf{CR} & \textbf{CF1} \\
  \midrule
  AP & $56.1$ & $78.5$ & $51.1$ & $61.9$ & $64.7$ & $40.9$ & $50.1$ \\
  FP & $73.9$ & $80.1$ & $67.8$ & $73.4$ & $78.2$ & $60.5$ & $68.2$ \\
  \midrule
  CP & $\mathbf{79.1}$ & $\mathbf{84.5}$ & $\mathbf{72.6}$ & $\mathbf{78.1}$ & $\mathbf{81.4}$ & $\mathbf{67.5}$ & $\mathbf{73.8}$ \\
  \bottomrule
  \end{tabular}
\end{table}

\textbf{Ablation Study for Semantic-aware Compact Prediction.}\label{sec:ablation-compact-prediciton} The final output from the interactive decoder is constructed of semantic tensor groups, where we use independent networks $\tau_1, \tau_2, \dots, \tau_C$ to predict the existence of labels. We conduct an ablation study to the compact prediction by replacing it with other solutions:
\begin{itemize}
  \item The semantic-aware compact prediction is denoted \texttt{CP}.
  \item For the average prediction (\texttt{AP}), we replace the original with a fully connected layer $\tau_{\mathrm{AP}}: \mathbb R^{D} \mapsto \mathbb R^C$, which has the same amount of parameters as \texttt{CP} does, and perform the following process to predict:
  \begin{equation}
    \mathbf p' = \mathtt{Mean}(\tau_{\mathrm{AP}}(\mathbf O_1,:),\tau_{\mathrm{AP}}(\mathbf O_2,:),\dots,\tau_{\mathrm{AP}}(\mathbf O_C,:)).
  \end{equation}
  \item For the fully-connected prediction (\texttt{FP}), we replace the original with a fully connected layer $\tau_{\mathrm{FP}}: \mathbb R^{CD} \mapsto \mathbb R^C$. $\{\mathbf O_c\}_{c=1}^C$ are flattened and concatenated into $\mathbf O \in \mathbb R^{CD}$ as the input for $\tau_{\mathrm{FP}}$.
\end{itemize}
We report the results in \Cref{tab:ablation-prediction}. \texttt{AP} show a catastrophic decrease compared with \texttt{CP}, while they have the same parameters. \texttt{FP} obtains a better result than \texttt{AP} does, but it is still lower than \texttt{CP} though \texttt{FP} has more parameters than \texttt{CP}. The results demonstrate that semantic-aware compact predictions regularize the representations in a generalized space, and significantly outperform other common solutions, \ie, \texttt{AP} and \texttt{FP}, with the least parameters. This also proves to us once again that more parameters do not always lead to better performance, which is even more evident in this task because of the overfitting brought by the lack of annotations.

\begin{table*}[t]
  \caption{Different evaluation metrics (OP, OR, OF1, CP, CR, CF1) on the PASCAL VOC \cite{pascal-voc-2007}, MS-COCO \cite{coco}, and VG-200 \cite{krishna2017visual} benchmarks.}
  \label{tab:evaluation-metrics}
  \centering
  \setlength{\tabcolsep}{3.7mm}
  \begin{tabular}{c|c||cccccc}
  \toprule
  \textbf{Benchmarks} & \textbf{Methods} & \textbf{OP} & \textbf{OR} & \textbf{OF1} & \textbf{CP} & \textbf{CR} & \textbf{CF1}\\
  \midrule
  \multirow{9}{*}{MS-COCO \cite{coco}} & SSGRL \cite{Chen2019SSGRL} & $86.3$ & $64.8$ & $73.9$ & $82.1$ & $58.4$ & $68.1$  \\
   & GCN-ML \cite{Chen2019GCNML} & $85.2$ & $64.2$ & $73.1$ & $81.8$ & $58.9$ & $68.4$\\
   & KGGR \cite{Chen2022KGGR} & $84.0$ & $65.6$ & $73.7$ & $81.4$ & $60.9$ & $69.7$ \\
   & Curriculum Learning \cite{Durand_2019_CVPR} & $\mathbf{87.8}$ & $51.0$ & $61.9$ & $60.9$ & $40.4$ & $48.3$ \\
   & Partial BCE \cite{Durand_2019_CVPR} & $\underline{86.7}$ & $64.7$ & $74.0$ & $\underline{83.1}$ & $58.9$ & $68.8$ \\
   & SST \cite{sst_aaai2022} & $86.3$ & $67.7$ & $75.8$ & $82.8$ & $62.6$ & $71.2$ \\
   & SARB \cite{Pu2022SARB} & $86.6$ & $68.6$ & $76.5$ & $82.9$ & $64.1$ & $72.2$ \\
   & DualCoOp* \cite{sun2022dualcoop} & $72.3$ & $\mathbf{81.5}$ & $\underline{76.6}$ & $70.1$ & $\mathbf{77.3}$ & $\underline{73.0}$ \\
   & Ours & $\mathbf{94.2}$ & $\mathbf{89.0}$ & $\mathbf{91.5}$ & $\mathbf{92.7}$ & $\mathbf{87.0}$ & $\mathbf{89.7}$ \\
  \midrule
  \multirow{9}{*}{PASCAL VOC \cite{pascal-voc-2007}} & SSGRL \cite{Chen2019SSGRL} & $91.2$ & $84.4$ & $87.7$ & $87.8$ & $81.4$ & $84.5$ \\
  & GCN-ML \cite{Chen2019GCNML} & $92.2$ & $83.0$ & $87.3$ & $89.7$ & $80.1$ & $84.6$\\
  & KGGR \cite{Chen2022KGGR} & $90.5$ & $82.9$ & $86.5$ & $88.5$ & $81.4$ & $84.7$\\
  & Curriculum Learning \cite{Durand_2019_CVPR} & $92.7$ & $78.2$ & $83.8$ & $79.5$ & $71.7$ & $75.4$ \\
  & Partial BCE \cite{Durand_2019_CVPR} & $91.8$ & $84.3$ & $87.9$ & $88.8$ & $81.3$ & $84.8$ \\
  & SST \cite{sst_aaai2022} & $91.3$ & $\underline{85.3}$ & $88.2$ & $88.3$ & $\underline{83.0}$ & $85.6$ \\
  & SARB \cite{Pu2022SARB} & $\underline{93.0}$ & $83.6$ & $\underline{88.4}$ & $\underline{90.4}$ & $81.1$ & $\underline{85.9}$\\
  & Ours & $\mathbf{94.2}$ & $\mathbf{89.0}$ & $\mathbf{91.5}$ & $\mathbf{92.7}$ & $\mathbf{87.0}$ & $\mathbf{89.7}$ \\
  \midrule
  \multirow{8}{*}{VG-200 \cite{krishna2017visual}} & SSGRL \cite{Chen2019SSGRL} & 69.9 & 25.9 & 37.8 & 45.3 & 18.3 & 26.1 \\
    & GCN-ML \cite{Chen2019GCNML} & 64.1 & 28.2 & 38.7 & 44.6 & 18.2 & 25.6 \\
    & KGGR \cite{Chen2022KGGR} & 64.5 & 30.5 & 41.2 & 54.8 & 25.8 & 33.6 \\
    & Curriculum labeling \cite{Durand_2019_CVPR} & 66.4 & 15.4 & 23.6 & 20.4 & 7.6 & 10.9 \\
    & partial-BCE \cite{Durand_2019_CVPR} & 69.7 & 24.6 & 36.1 & 44.3 & 18.1 & 25.7 \\
    & SST \cite{sst_aaai2022} & 69.9 & 27.9 & 39.9 & 49.8 & 22.3 & 30.8 \\
    & SARB \cite{Pu2022SARB} & $\underline{70.1}$ & $\underline{33.2}$ & $\underline{45.0}$ & $\underline{56.8}$ & $\underline{27.8}$ & $\underline{37.4}$ \\
    & Ours & $\mathbf{73.7}$& $\mathbf{42.7}$& $\mathbf{54.0}$& $\mathbf{65.1}$& $\mathbf{38.6}$& $\mathbf{48.4}$ \\
  \bottomrule
  \multicolumn{8}{l}{*: Reproduced under the same data partition.}
  \end{tabular}
\end{table*}

\begin{table}[t]
  \centering
  \caption{The statistics for the inference time of ours and three state-of-the-art methods on the MS-COCO dataset.}\label{tab:time-params-stat}
  \setlength{\tabcolsep}{3.7mm}
  \begin{tabular}{c|cc}
  \toprule
  \textbf{Methods} & \textbf{Inference Time} & \textbf{Avg. mAP} \\
  \midrule
  SST \cite{sst_aaai2022} & $20~\mathrm{ms}~\textcolor{teal}{(-\phantom{0}3)}$ & $76.7~\textcolor{blue}{(-6.2)}$ \\
  SARB \cite{Pu2022SARB} & $19~\mathrm{ms}~\textcolor{teal}{(-\phantom{0}4)}$ & $77.9~\textcolor{blue}{(-5.0)}$ \\
  DualCoOp \cite{sun2022dualcoop} & $62~\mathrm{ms}~\textcolor{blue}{(+39)}$ & $79.2~\textcolor{blue}{(-3.7)}$ \\
  \midrule
  Ours & $23~\mathrm{ms}~\phantom{(+00)}$ & $82.9~\phantom{(+0.0)}$ \\
  \bottomrule
  \end{tabular}
\end{table}

\textbf{Time Efficiency and Performance Evaluation.}\label{sec:time-efficiency}
To analyze the efficiency of methods, we conduct experiments to compare the inference time between ours and three state-of-the-art methods, SST \cite{sst_aaai2022}, SARB \cite{Pu2022SARB}, and DualCoOp \cite{sun2022dualcoop}. The results are reported in \Cref{tab:time-params-stat}.
In detail, we resize the input to resolution $448 \times 448$ and run all methods for $2,000$ batches to count the average inference time on a single NVIDIA 3090 GPU. Due to the introduction of the text, language-vision multi-modal methods tend to consume more inference time. From \Cref{tab:time-params-stat}, the inference time of DualCoOp is much longer than that of other vision-only methods, \ie, SST and SARB. By optimizing the implementation to cache the language prototypes, our method eliminates the text encoder and can reduce the inference time significantly compared with vanilla CLIP, achieving better performance while keeping a balanced time consumption. In summary, our method not only trains fast but also is resource-efficient while obtaining the best performance in inference.

\section{Discussion and Interpretation} \label{sec:interpre}

\begin{figure}[t]
  \centering
  \includegraphics[width=\linewidth]{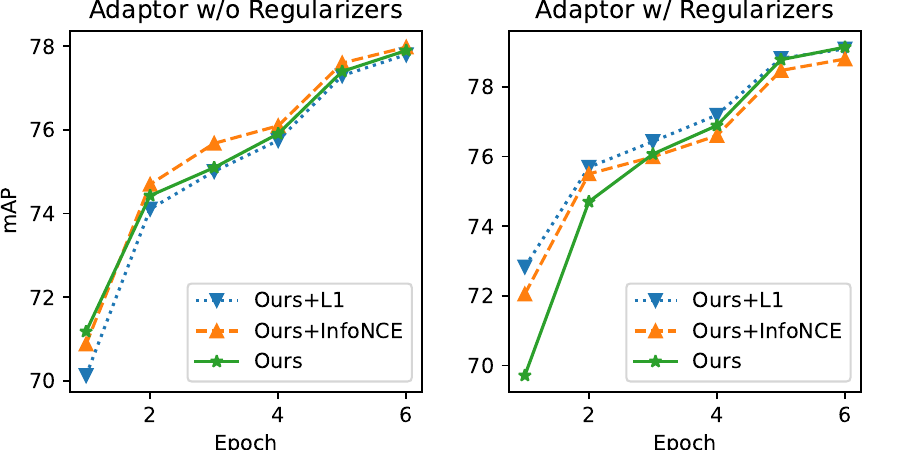}
  \caption{The mAP during training under different contrastive configurations.}
  \label{fig:cl-convergence}
\end{figure}

\subsection{How Does Implicit Contrastive Learning Work?} To verify the effects of implicit contrastive learning, we add the proposed implicit contrastive learning (denoted as Implicit) with InfoNCE and positive L1 losses in \Cref{sec:loss}.  From \Cref{tab:ablation-effect} and \Cref{fig:cl-convergence}, we observed that: 

\begin{enumerate}
    \item InfoNCE and L1 constraints play a similar role with our implicit constraints, as adding these constraints shows a fast convergence speed but similar final results.
    \item Negative training without semantic constraints makes the performance worse,~\ie, from $79.1$ to $78.7$. Compared with positive pairs, image-level negative pairs corrupt the model because of the semantic inconsistency in multi-label learning.
    \item Solely enhancing the positive constraints by L1 loss shows slight variations with the implicit model, indicating our proposed semantic contrastive learning covers the positive relationships. To sum up, our implicit contrastive learning both considers the positive and negative relationships, achieving better results than adding additional image-level contrastive constraints.
\end{enumerate}

\begin{figure*}[t]
  \centering
  \includegraphics[width=\linewidth]{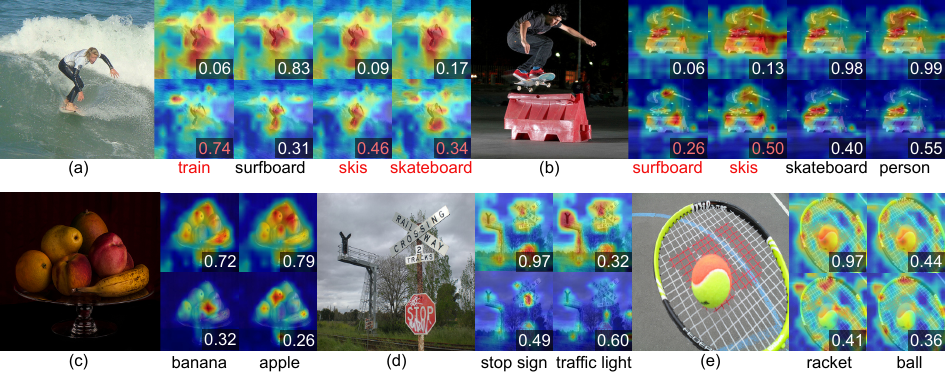}
  \caption{Visualization of class activation maps on MS-COCO \cite{coco} via GradCAM++ \cite{Chattopadhay2018GradCAMplusplus}. CAMs in the first and second lines of each group are exported from our LDSA and the baseline respectively. The confidences predicted by models are labeled on CAMs. The labels that are not present in images are colored red.}\label{fig:cam}
\end{figure*}

\begin{figure}[t]
  \centering
  \includegraphics[width=\linewidth]{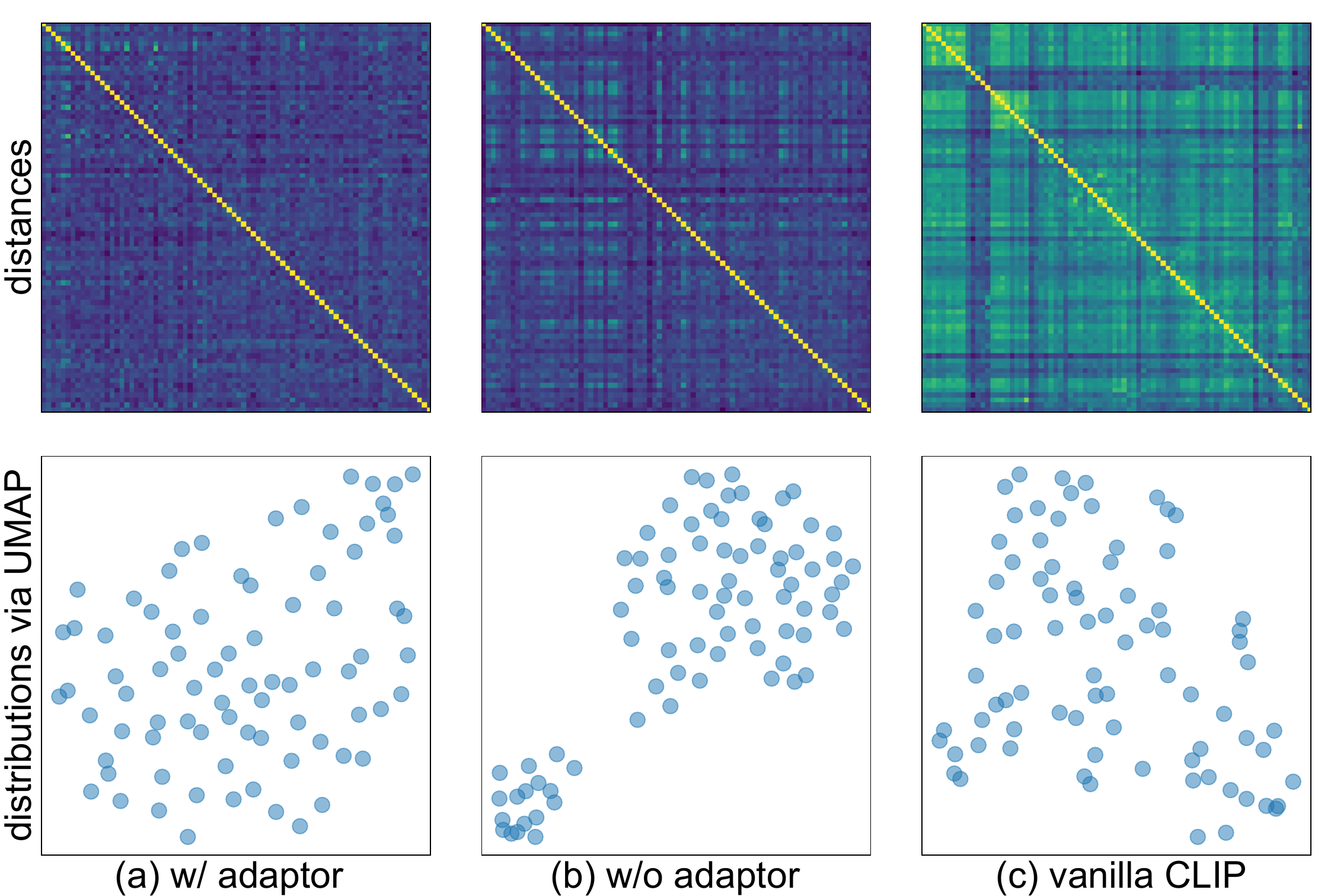}
  \caption{The visualization of distances and distributions of language prototypes.}
  \label{fig:prompts-distribution}
\end{figure}

\begin{figure}[t]
  \centering
  \includegraphics[width=.965\linewidth]{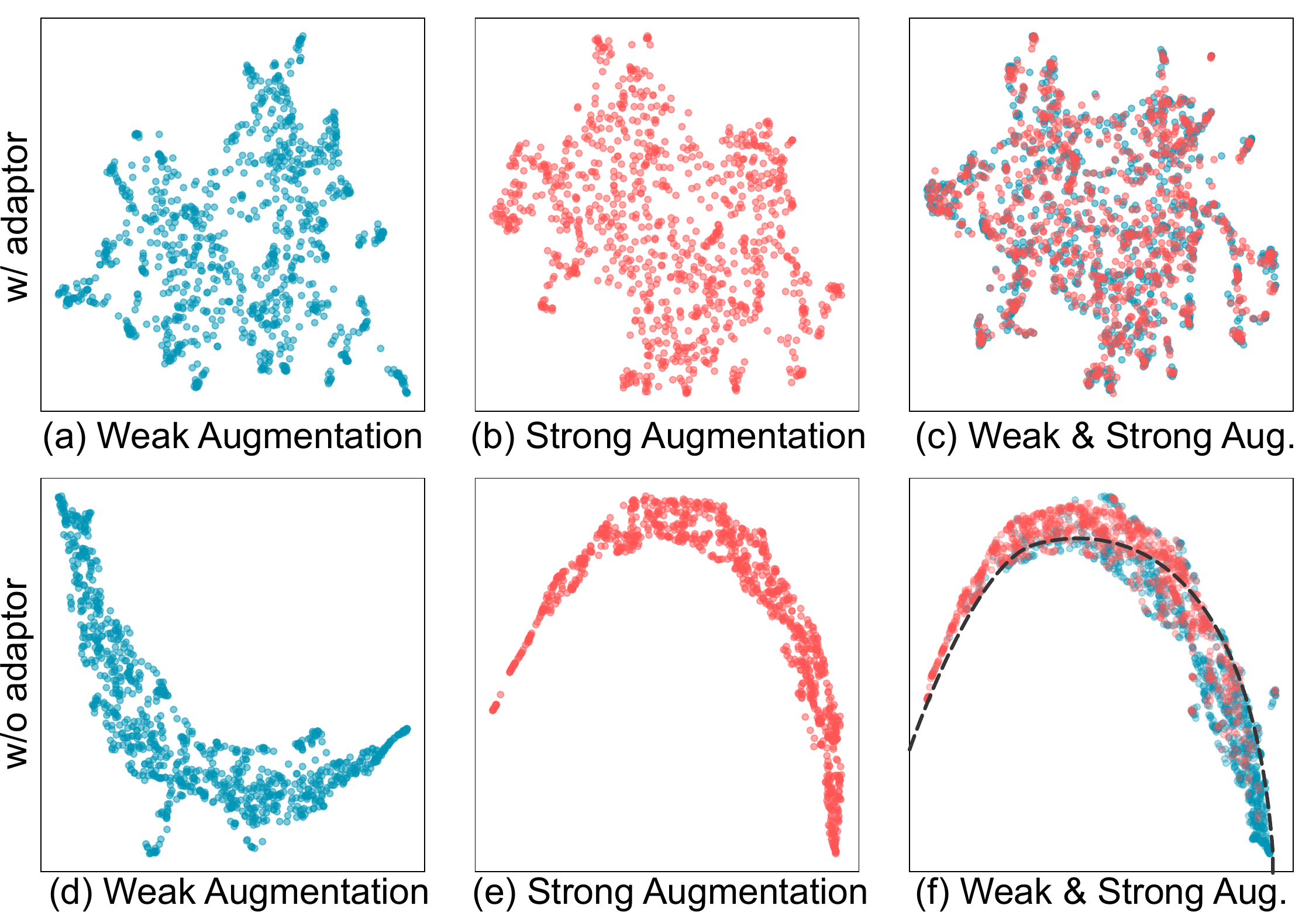}
  \caption{The feature manifolds of augmented images. Our model in c) clusters samples by its semantics while the ablated model w/o adaptor focuses on the visual appearances with a clear decision boundary in f).}
  \label{fig:image-distribution}
\end{figure}

\subsection{Contrastive Adaptor Helps Visual Feature Generalization}  We visualize the distribution of weakly and strongly augmented features of random $1,000$ images from MS-COCO via UMAP~\cite{mcinnes2018umap-software}. The features encoded w/ and w/o the adaptor are plotted in \Cref{fig:image-distribution}. The weak and strong augmentation in the third row exhibits the joint distribution in feature space. The proposed contrastive adaptor forms weak and strong augmentations in a uniformly distributed space. While in the joint distributions in \Cref{fig:image-distribution} c) and f), models w/o the adaptor show a clear decision boundary in this low-dimensional space, while our full model in c) clusters different views of augmentation only by its semantics.

\subsection{Contrastive Adaptor Helps The Generalization of Language Prototypes} We visualize the distances and distributions of language prototypes in \Cref{fig:prompts-distribution}. It shows prototype distribution with adaptor is distributed as uniform as those by vanilla CLIP, while the former is distributed in a distant space. Although prototypes w/o the adaptor are distinguishable in distance, the information in them is partly lost according to their distribution. These results demonstrate that our model transfers the semantic knowledge to the downstream datasets without losing generalization.

\begin{table}[t]
  \caption{Ablation study for contrastive effects on MS-COCO. CL: Contrastive learning. Regularizers: Dense feature regularizers.}
  \label{tab:ablation-effect}
  \centering
  \begin{tabular}{c||c|c}
    \toprule
    \textbf{CL} & \textbf{w/o Regularizers} & \textbf{w/  Regularizers} \\
    \midrule
    Ours Implicit & $77.9\textcolor{black}{(+0.0)}$ & $79.1\textcolor{black}{(+0.0)}$ \\
    + InfoNCE & $78.0\textcolor{red}{(+0.1)}$ & $78.7\textcolor{blue}{(-0.4)}$ \\
    + L1 Loss& $77.8\textcolor{blue}{(-0.1)}$ & $79.1\textcolor{black}{(+0.0)}$ \\
    \bottomrule
    \end{tabular}
\end{table}

\subsection{Where does LDSA Focus?}\label{sec:visulization}

One crucial idea of our proposed LDSA is to excavate the adaptive semantic relationships to enhance multi-label classification with partial annotations. Besides the visualization from the textual side in \Cref{fig:prompts-distribution}, here we visualize the network activations to images in \Cref{fig:cam} via GradCAM++ \cite{Chattopadhay2018GradCAMplusplus}, to analyze it in terms of vision. First, the CAMs of our proposed methods are more precise. LDSA pays attention to the \textit{person, banana, traffic lights} in \Cref{fig:cam} b) c) d), while the baseline overfits to the \textit{skateboard, apple}, and even background noise improperly. Second, LDSA has an awareness of the scenes and object interactions. For example, in \Cref{fig:cam} a) and b) LDSA not only ``\textbf{focuses}'' on the surfboard and skateboard, masked with red color to show its importance, but also ``\textbf{sees}'' the person and the surrounding background. The baseline limits its fields of view and drastically fails to tell them from each other in those images. Since the \textit{skateboard} and \textit{surfboard} are similar in vision, how to distinguish them is mainly based on semantic information hiding in the interactions with other objects and the scenes, which can be captured by LDSA on its own. Such ability helps LDSA improve its performance.

\subsection{Semantic Relations by Different Prior Assumptions} 
\begin{figure}[t]
  \centering
  \includegraphics[width=\linewidth]{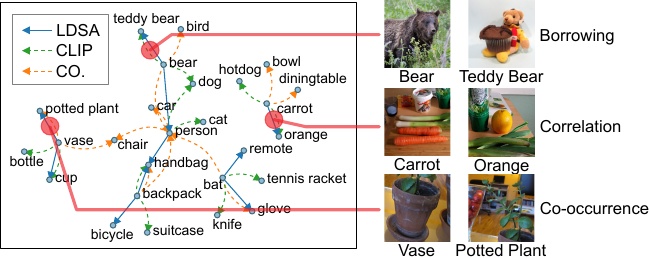}
  \caption{Semantic relationships of representative labels on MS-COCO. The labels are connected by the cosine similarity of our language prototypes (\textbf{LDSA}), label CO-occurrence exported from the training set (\textbf{CO.}), and cosine similarity of encoded prompts by vanilla CLIP (\textbf{CLIP}), respectively. We highlight three edges and analyze their relationships.}
  \label{fig:semantic-graph}
\end{figure}

What semantic relationships do we learn? To verify this, we exhibit the semantic relations on the MS-COCO dataset in \Cref{fig:semantic-graph} with three prior assumptions: 
\begin{enumerate}
    \item Label Co-occurrence (CO.) learns pre-determined relationships while some relationships are not feasible in the downstream task.
    \item CLIP learns generic relationships but shows weak adaptation to the downstream task.
    \item Our proposed LDSA learns \textbf{prior-adaptive} correlations which not only select the meaningful co-occurrence (\textit{vase} and \textit{plotted plant}) but also with other implicit relationships,~\eg,  semantic meaning borrowing (\textit{bear} and \textit{teddy bear}), coarse semantic correlation  (\textit{carrot} and \textit{orange} with similar color).
\end{enumerate}
Instead of using fixed, fragile semantic relationships, our LDSA learns to excavate adaptive generalized knowledge on its own, thus achieving significant performance improvement.

\subsection{Few-shot Generalization}

\begin{table}[t]
  \caption{Few-shot experimental comparison using mAP (\%) on the MS-COCO dataset.}
  \label{tab:few-shot}
  \centering
  \begin{tabular}{c|cc}
  \toprule
  \textbf{Methods} & 1-shot & 5-shot\\
  \midrule
  LaSO \cite{Alfassy2019laso} & 45.3 & 58.1 \\
  ML-FSL \cite{Simon2022mlfsl} & $\underline{54.4}$ & $\mathbf{63.6}$ \\
  CoOp \cite{zhou2022coop} & 46.9 & 55.6 \\
  Tip-Adapter \cite{zhang2021tipadaptertrainingfreeclipadapterbetter} & 53.8 & 59.7 \\
  \midrule
  Ours & $\mathbf{56.3}$ & $\underline{60.0}$ \\
  \bottomrule
  \end{tabular}
\end{table}

To further examine the generalization behavior of LDSA under limited supervision, we evaluate it in the few-shot setting on MS-COCO. As shown in \Cref{tab:few-shot}, LDSA achieves the best result in the 1-shot setting and remains competitive in the 5-shot setting. These results suggest that the proposed task-specific adaptation is effective for partial-label multi-label learning while retaining promising generalization ability with limited supervision.

\subsection{Limitations and Future Works}

While our method significantly enhances performance through class-specific modules that boost partial-label accuracy, we recognize that this specialization may limit open-set generalization compared to more generic prompt learning methods.
Although we have analyzed the underlying contrastive effects through extensive experiments and visualizations, our exploration remains at a preliminary stage. Future efforts should focus on bridging this gap by seeking a deeper theoretical understanding of these principles, aiming to reconcile high-precision specialization with broader generalizability across diverse domains.

\section{Conclusions}\label{sec:conclusion}
To improve prior assumptions on multi-label learning tasks, in this paper, we propose Language-driven Dense Semantic Adaptor (LDSA) for multi-label classification with partial annotations. LDSA excavates prior-adaptive generalized knowledge from two aspects,~\ie, a densely contrastive adaptor for visual representation and a language-driven interactive decoder to transfer the generalized language knowledge to visual domains.  With the joint learning of visual and language representations, our proposed approach achieves state-of-the-art on the MS-COCO, PASCAL VOC and Visual Genome datasets for this partial multi-label learning task.

\ifCLASSOPTIONcaptionsoff
  \newpage
\fi

\balance
\bibliographystyle{IEEEtran}
\bibliography{incfewshot}

@string(CVPR=		{IEEE/CVF Conference on Computer Vision and Pattern Recognition (CVPR)})

@string(ICCV=		{IEEE/CVF International Conference on Computer Vision (ICCV)})

@string(NIPS=   {Advances in Neural Information Processing Systems (NeurIPS)})

@string(ECCV=		{European Conference on Computer Vision (ECCV)})

@string(AAAI=		{AAAI Conference on Artificial Intelligence (AAAI)})

@string(IJCV	=		{International Journal of Computer Vision})

@string(PAMI	=		{IEEE Transactions on Pattern Analysis and Machine Intelligence})

@string(TIP		=		{IEEE Transactions on Image Processing})

@string(TMM		=		{IEEE Transactions on Multimedia})

@string(ICML		=	{International Conference on Machine Learning (ICML)}	)

@article{gcn_tpami,
  author  = {Chen, Zhaomin and Wei, Xiu-Shen and Wang, Peng and Guo, Yanwen},
  journal = PAMI,
  title   = {Learning Graph Convolutional Networks for Multi-Label Recognition and Applications},
  year    = {2021},
  volume  = {},
  number  = {},
  pages   = {1-1},
  doi     = {10.1109/TPAMI.2021.3063496}
}

@article{sst_tip,
  author  = {Chen, Zhao-Min and Cui, Quan and Zhao, Borui and Song, Renjie and Zhang, Xiaoqin and Yoshie, Osamu},
  journal = TIP,
  title   = {SST: Spatial and Semantic Transformers for Multi-Label Image Recognition},
  year    = {2022},
  volume  = {31},
  number  = {},
  pages   = {2570-2583},
  doi     = {10.1109/TIP.2022.3148867}
}

@article{mcra_tip,
  author  = {Gao, Bin-Bin and Zhou, Hong-Yu},
  journal = TIP,
  title   = {Learning to Discover Multi-Class Attentional Regions for Multi-Label Image Recognition},
  year    = {2021},
  volume  = {30},
  number  = {},
  pages   = {5920-5932},
  doi     = {10.1109/TIP.2021.3088605}
}

@inproceedings{deng2009imagenet,
  title     = {Imagenet: A large-scale hierarchical image database},
  author    = {Deng, Jia and Dong, Wei and Socher, Richard and Li, Li-Jia and Li, Kai and Fei-Fei, Li},
  booktitle = CVPR,
  year      = {2009}
}

@inproceedings{coco,
  author    = {Lin, Tsung-Yi
               and Maire, Michael
               and Belongie, Serge
               and Hays, James
               and Perona, Pietro
               and Ramanan, Deva
               and Doll{\'a}r, Piotr
               and Zitnick, C. Lawrence},
  title     = {Microsoft COCO: Common Objects in Context},
  booktitle = ECCV,
  year      = {2014}
}

@article{sst_aaai2022,
  doi      = {10.48550/ARXIV.2112.10941},
  author   = {Chen, Tianshui and Pu, Tao and Wu, Hefeng and Xie, Yuan and Lin, Liang},
  title    = {Structured Semantic Transfer for Multi-Label Recognition with Partial Labels},
  journal  = {arXiv preprint arXiv:2112.10941},
  year     = {2021}
}

@inproceedings{Pu2022SARB,
  title     = {Semantic-Aware Representation Blending for Multi-Label Image Recognition with Partial Labels},
  author    = {Pu, Tao and Chen, Tianshui and Wu, Hefeng and Lin, Liang},
  booktitle = AAAI,
  year      = {2022}
}

@inproceedings{Durand_2019_CVPR,
  author    = {Durand, Thibaut and Mehrasa, Nazanin and Mori, Greg},
  title     = {Learning a Deep ConvNet for Multi-Label Classification With Partial Labels},
  booktitle = CVPR,
  year      = {2019}
}

@inproceedings{interactiveCNN_CVPR,
  author    = {Huynh, Dat and Elhamifar, Ehsan},
  title     = {Interactive Multi-Label CNN Learning With Partial Labels},
  booktitle = CVPR,
  year      = {2020}
}

@inproceedings{Rajeswar_2022_CVPR,
  author    = {Rajeswar, Sai and Rodr{\'\i}guez, Pau and Singhal, Soumye and Vazquez, David and Courville, Aaron},
  title     = {Multi-Label Iterated Learning for Image Classification With Label Ambiguity},
  booktitle = CVPR,
  month     = {June},
  year      = {2022},
  pages     = {4783-4793}
}

@inproceedings{Ben-Baruch_2022_CVPR,
  author    = {Ben-Baruch, Emanuel and Ridnik, Tal and Friedman, Itamar and Ben-Cohen, Avi and Zamir, Nadav and Noy, Asaf and Zelnik-Manor, Lihi},
  title     = {Multi-Label Classification With Partial Annotations Using Class-Aware Selective Loss},
  booktitle = CVPR,
  month     = {June},
  year      = {2022},
  pages     = {4764-4772}
}

@article{Sun2022globallocal,
  author  = {Sun, Lijuan and Feng, Songhe and Liu, Jun and Lyu, Gengyu and Lang, Congyan},
  journal = TMM,
  title   = {Global-Local Label Correlation for Partial Multi-Label Learning},
  year    = {2022},
  volume  = {24},
  number  = {},
  pages   = {581-593},
  doi     = {10.1109/TMM.2021.3055959}
}

@inproceedings{mlgcn_CVPR2019,
  author    = {Chen, Zhao-Min and Wei, Xiu-Shen and Wang, Peng and Guo, Yanwen},
  title     = {Multi-Label Image Recognition With Graph Convolutional Networks},
  booktitle = CVPR,
  year      = {2019}
}

@article{PiCO,
  doi      = {10.48550/ARXIV.2201.08984},
  author   = {Wang, Haobo and Xiao, Ruixuan and Li, Yixuan and Feng, Lei and Niu, Gang and Chen, Gang and Zhao, Junbo},
  title    = {PiCO: Contrastive Label Disambiguation for Partial Label Learning},
  journal  = {arXiv preprint arXiv:2201.08984},
  year     = {2022}
}

@article{Tsoumakas_2009_MLOverview,
  author  = {Tsoumakas, Grigorios and Katakis, Ioannis},
  year    = {2009},
  title   = {Multi-Label Classification: An Overview},
  journal = {International Journal of Data Warehousing and Mining},
  doi     = {10.4018/jdwm.2007070101}
}

@inproceedings{role_CVPR2021,
  author    = {Cole, Elijah and Mac Aodha, Oisin and Lorieul, Titouan and Perona, Pietro and Morris, Dan and Jojic, Nebojsa},
  title     = {Multi-Label Learning From Single Positive Labels},
  booktitle = CVPR,
  year      = {2021}
}

@inproceedings{tdrg_ICCV2021,
  author    = {Zhao, Jiawei and Yan, Ke and Zhao, Yifan and Guo, Xiaowei and Huang, Feiyue and Li, Jia},
  title     = {Transformer-Based Dual Relation Graph for Multi-Label Image Recognition},
  booktitle = ICCV,
  year      = {2021}
}

@article{corruptimagenet_2019,
  doi       = {10.1109/access.2019.2956775},
  year      = 2019,
  publisher = {Institute of Electrical and Electronics Engineers ({IEEE})},
  author    = {Baoyuan Wu and Weidong Chen and Yanbo Fan and Yong Zhang and Jinlong Hou and Jie Liu and Tong Zhang},
  title     = {Tencent {ML}-Images: A Large-Scale Multi-Label Image Database for Visual Representation Learning},
  journal   = {{IEEE} Access}
}

@InProceedings{univip,
    author    = {Li, Zhaowen and Zhu, Yousong and Yang, Fan and Li, Wei and Zhao, Chaoyang and Chen, Yingying and Chen, Zhiyang and Xie, Jiahao and Wu, Liwei and Zhao, Rui and Tang, Ming and Wang, Jinqiao},
    title     = {UniVIP: A Unified Framework for Self-Supervised Visual Pre-Training},
    booktitle = {Proceedings of the IEEE/CVF Conference on Computer Vision and Pattern Recognition (CVPR)},
    month     = {June},
    year      = {2022},
    pages     = {14627-14636}
}

@inproceedings{byol,
  author    = {Grill, Jean-Bastien and Strub, Florian and Altch\'{e}, Florent and Tallec, Corentin and Richemond, Pierre and Buchatskaya, Elena and Doersch, Carl and Avila Pires, Bernardo and Guo, Zhaohan and Gheshlaghi Azar, Mohammad and Piot, Bilal and kavukcuoglu, koray and Munos, Remi and Valko, Michal},
  booktitle = NIPS,
  title     = {Bootstrap Your Own Latent - A New Approach to Self-Supervised Learning},
  year      = {2020}
}

@inproceedings{moco_CVPR2020,
  author    = {He, Kaiming and Fan, Haoqi and Wu, Yuxin and Xie, Saining and Girshick, Ross},
  title     = {Momentum Contrast for Unsupervised Visual Representation Learning},
  booktitle = CVPR,
  year      = {2020}
}

@inproceedings{simclr,
  title     = {A Simple Framework for Contrastive Learning of Visual Representations},
  author    = {Chen, Ting and Kornblith, Simon and Norouzi, Mohammad and Hinton, Geoffrey},
  booktitle = ICML,
  year      = {2020}
}

@misc{clip,
  doi       = {10.48550/ARXIV.2103.00020},
  url       = {https://arxiv.org/abs/2103.00020},
  author    = {Radford, Alec and Kim, Jong Wook and Hallacy, Chris and Ramesh, Aditya and Goh, Gabriel and Agarwal, Sandhini and Sastry, Girish and Askell, Amanda and Mishkin, Pamela and Clark, Jack and Krueger, Gretchen and Sutskever, Ilya},
  title     = {Learning Transferable Visual Models From Natural Language Supervision},
  publisher = {arXiv},
  year      = {2021},
  copyright = {arXiv.org perpetual, non-exclusive license}
}

@article{zhou2022coop,
  title   = {Learning to Prompt for Vision-Language Models},
  author  = {Zhou, Kaiyang and Yang, Jingkang and Loy, Chen Change and Liu, Ziwei},
  journal = IJCV,
  year    = {2022}
}

@article{gao2021clipadapter,
  title   = {CLIP-Adapter: Better Vision-Language Models with Feature Adapters},
  author  = {Gao, Peng and Geng, Shijie and Zhang, Renrui and Ma, Teli and Fang, Rongyao and Zhang, Yongfeng and Li, Hongsheng and Qiao, Yu},
  journal = {arXiv preprint arXiv:2110.04544},
  year    = {2021}
}

@inproceedings{sun2022dualcoop,
  title     = {DualCoOp: Fast Adaptation to Multi-Label Recognition with Limited Annotations},
  author    = {Ximeng Sun and Ping Hu and Kate Saenko},
  booktitle = NIPS,
  editor    = {Alice H. Oh and Alekh Agarwal and Danielle Belgrave and Kyunghyun Cho},
  year      = {2022},
  url       = {https://openreview.net/forum?id=QnajmHkhegH}
}

@misc{pascal-voc-2007,
  author       = {Everingham, M. and Van~Gool, L. and Williams, C. K. I. and Winn, J. and Zisserman, A.},
  title        = {The {PASCAL} {V}isual {O}bject {C}lasses {C}hallenge 2007 {(VOC2007)} {R}esults},
  howpublished = {http://www.pascal-network.org/challenges/VOC/voc2007/workshop/index.html},
  year         = {2007}
}

@inproceedings{Ridnik2021asymmetricloss,
  author    = {Ridnik, Tal and Ben-Baruch, Emanuel and Zamir, Nadav and Noy, Asaf and Friedman, Itamar and Protter, Matan and Zelnik-Manor, Lihi},
  title     = {Asymmetric Loss for Multi-Label Classification},
  booktitle = ICCV,
  month     = {October},
  year      = {2021},
  pages     = {82-91}
}

@inproceedings{Chen2019SSGRL,
  author    = {Chen, Tianshui and Xu, Muxin and Hui, Xiaolu and Wu, Hefeng and Lin, Liang},
  title     = {Learning Semantic-Specific Graph Representation for Multi-Label Image Recognition},
  booktitle = ICCV,
  year      = {2019}
}

@inproceedings{JIANG2018mentornet,
  title     = {{M}entor{N}et: Learning Data-Driven Curriculum for Very Deep Neural Networks on Corrupted Labels},
  author    = {Jiang, Lu and Zhou, Zhengyuan and Leung, Thomas and Li, Li-Jia and Fei-Fei, Li},
  booktitle = ICML,
  year      = {2018}
}

@inproceedings{Medhini2021clipit,
  author    = {Narasimhan, Medhini and Rohrbach, Anna and Darrell, Trevor},
  booktitle = NIPS,
  editor    = {M. Ranzato and A. Beygelzimer and Y. Dauphin and P.S. Liang and J. Wortman Vaughan},
  pages     = {13988--14000},
  publisher = {Curran Associates, Inc.},
  title     = {CLIP-It! Language-Guided Video Summarization},
  url       = {https://proceedings.neurips.cc/paper/2021/file/7503cfacd12053d309b6bed5c89de212-Paper.pdf},
  volume    = {34},
  year      = {2021}
}

@inproceedings{Patashnik2021styleCLIP,
  author    = {Patashnik, Or and Wu, Zongze and Shechtman, Eli and Cohen-Or, Daniel and Lischinski, Dani},
  booktitle = ICCV,
  title     = {StyleCLIP: Text-Driven Manipulation of StyleGAN Imagery},
  year      = {2021},
  volume    = {},
  number    = {},
  pages     = {2065-2074},
  doi       = {10.1109/ICCV48922.2021.00209}
}

@inproceedings{Kim2022diffusionCLIP,
  author    = {Kim, Gwanghyun and Kwon, Taesung and Ye, Jong Chul},
  title     = {DiffusionCLIP: Text-Guided Diffusion Models for Robust Image Manipulation},
  booktitle = CVPR,
  month     = {June},
  year      = {2022},
  pages     = {2426-2435}
}

@inproceedings{Wang2022CRIS,
  author    = {Wang, Zhaoqing and Lu, Yu and Li, Qiang and Tao, Xunqiang and Guo, Yandong and Gong, Mingming and Liu, Tongliang},
  title     = {CRIS: CLIP-Driven Referring Image Segmentation},
  booktitle = CVPR,
  month     = {June},
  year      = {2022},
  pages     = {11686-11695}
}

@inproceedings{Wang2022clipNerf,
  author    = {Wang, Can and Chai, Menglei and He, Mingming and Chen, Dongdong and Liao, Jing},
  title     = {CLIP-NeRF: Text-and-Image Driven Manipulation of Neural Radiance Fields},
  booktitle = CVPR,
  month     = {June},
  year      = {2022},
  pages     = {3835-3844}
}

@inproceedings{Ioffe2015batchNorm,
  author    = {Ioffe, Sergey and Szegedy, Christian},
  title     = {Batch Normalization: Accelerating Deep Network Training by Reducing Internal Covariate Shift},
  year      = {2015},
  publisher = {JMLR.org},
  booktitle = ICML,
  pages     = {448–456},
  numpages  = {9},
  location  = {Lille, France},
  series    = {ICML'15}
}

@misc{Ba2016layerNorm,
  doi       = {10.48550/ARXIV.1607.06450},
  url       = {https://arxiv.org/abs/1607.06450},
  author    = {Ba, Jimmy Lei and Kiros, Jamie Ryan and Hinton, Geoffrey E.},
  title     = {Layer Normalization},
  publisher = {arXiv},
  year      = {2016},
  copyright = {arXiv.org perpetual, non-exclusive license}
}

@inproceedings{Wu2018groupNorm,
  author    = {Wu, Yuxin and He, Kaiming},
  title     = {Group Normalization},
  booktitle = ECCV,
  month     = {September},
  year      = {2018}
}

@misc{bnInBYOL,
  author = {Abe, Fetterman and Josh, Albrecht},
  url    = {https://generallyintelligent.com/blog/2020-08-24-understanding-self-supervised-contrastive-learning/},
  title  = {Understanding Self-Supervised and Contrastive Learning with "Bootstrap Your Own Latent"},
  year   = {},
  note   = {Accessed: 2023-02-22}
}

@inproceedings{detr,
  author    = {Carion, Nicolas and Massa, Francisco and Synnaeve, Gabriel and Usunier, Nicolas and Kirillov, Alexander and Zagoruyko, Sergey},
  title     = {End-to-End Object Detection with Transformers},
  year      = {2020},
  doi       = {10.1007/978-3-030-58452-8\_13},
  booktitle = ECCV
}

@inproceedings{Chen2019GCNML,
  author    = {Chen, Zhao-Min and Wei, Xiu-Shen and Wang, Peng and Guo, Yanwen},
  title     = {Multi-Label Image Recognition With Graph Convolutional Networks},
  booktitle = CVPR,
  year      = {2019}
}

@article{Chen2022KGGR,
  author  = {Chen, Tianshui and Lin, Liang and Chen, Riquan and Hui, Xiaolu and Wu, Hefeng},
  journal = PAMI,
  title   = {Knowledge-Guided Multi-Label Few-Shot Learning for General Image Recognition},
  year    = {2022}
}

@article{mcinnes2018umap-software,
  title   = {UMAP: Uniform Manifold Approximation and Projection},
  author  = {McInnes, Leland and Healy, John and Saul, Nathaniel and Grossberger, Lukas},
  journal = {The Journal of Open Source Software},
  volume  = {3},
  number  = {29},
  pages   = {861},
  year    = {2018}
}

@misc{oord2018infonce,
  doi       = {10.48550/ARXIV.1807.03748},
  url       = {https://arxiv.org/abs/1807.03748},
  author    = {Oord, Aaron van den and Li, Yazhe and Vinyals, Oriol},
  title     = {Representation Learning with Contrastive Predictive Coding},
  publisher = {arXiv},
  year      = {2018},
  copyright = {arXiv.org perpetual, non-exclusive license}
}

@inproceedings{Kundu2020exploitweakly,
 author = {Kundu, Kaustav and Tighe, Joseph},
 booktitle = NIPS,
 editor = {H. Larochelle and M. Ranzato and R. Hadsell and M.F. Balcan and H. Lin},
 pages = {561--572},
 publisher = {Curran Associates, Inc.},
 title = {Exploiting weakly supervised visual patterns to learn from partial annotations},
 url = {https://proceedings.neurips.cc/paper/2020/file/066ca7bf90807fcd8e4f1eaef4e4e8f7-Paper.pdf},
 volume = {33},
 year = {2020}
}

@InProceedings{Wu2015MLMG,
author = {Wu, Baoyuan and Lyu, Siwei and Ghanem, Bernard},
title = {ML-MG: Multi-Label Learning With Missing Labels Using a Mixed Graph},
booktitle = ICCV,
month = {December},
year = {2015}
}

@InProceedings{Rao2022denseclip,
    author    = {Rao, Yongming and Zhao, Wenliang and Chen, Guangyi and Tang, Yansong and Zhu, Zheng and Huang, Guan and Zhou, Jie and Lu, Jiwen},
    title     = {DenseCLIP: Language-Guided Dense Prediction With Context-Aware Prompting},
    booktitle = CVPR,
    month     = {June},
    year      = {2022},
    pages     = {18082-18091}
}

@InProceedings{Zhou2022conditionalprompt,
    author    = {Zhou, Kaiyang and Yang, Jingkang and Loy, Chen Change and Liu, Ziwei},
    title     = {Conditional Prompt Learning for Vision-Language Models},
    booktitle = CVPR,
    month     = {June},
    year      = {2022},
    pages     = {16816-16825}
}

@INPROCEEDINGS{Chattopadhay2018gradcamplusplus,
  author={Chattopadhay, Aditya and Sarkar, Anirban and Howlader, Prantik and Balasubramanian, Vineeth N},
  booktitle={2018 IEEE Winter Conference on Applications of Computer Vision (WACV)}, 
  title={Grad-CAM++: Generalized Gradient-Based Visual Explanations for Deep Convolutional Networks}, 
  year={2018},
  volume={},
  number={},
  pages={839-847},
  doi={10.1109/WACV.2018.00097}}

@inproceedings{Guo2023textprompt,
  author    = {Guo, Zixian and Dong, Bowen and Ji, Zhilong and Bai, Jinfeng and Guo, Yiwen and Zuo, Wangmeng},
  title     = {Texts as Images in Prompt Tuning for Multi-Label Image Recognition},
  booktitle = CVPR,
  month     = {June},
  year      = {2023},
  pages     = {2808-2817}
}

@inproceedings{Zhu2023multiLabelSceneLearning,
  author    = {Zhu, Ke and Fu, Minghao and Wu, Jianxin},
  title     = {Multi-Label Self-Supervised Learning with Scene Images},
  booktitle = ICCV,
  month     = {October},
  year      = {2023},
  pages     = {6694-6703}
}

@inproceedings{Zhu2023sceneGraphMultiLabel,
  author    = {Zhu, Xuelin and Liu, Jian and Liu, Weijia and Ge, Jiawei and Liu, Bo and Cao, Jiuxin},
  title     = {Scene-Aware Label Graph Learning for Multi-Label Image Classification},
  booktitle = ICCV,
  month     = {October},
  year      = {2023},
  pages     = {1473-1482}
}

@inproceedings{Li2023CTMultiLabel,
  author    = {Li, Miaoge and Wang, Dongsheng and Liu, Xinyang and Zeng, Zequn and Lu, Ruiying and Chen, Bo and Zhou, Mingyuan},
  title     = {PatchCT: Aligning Patch Set and Label Set with Conditional Transport for Multi-Label Image Classification},
  booktitle = ICCV,
  month     = {October},
  year      = {2023},
  pages     = {15348-15358}
}

@inproceedings{Zhang2022MLContrastive,
  author    = {Zhang, Shu and Xu, Ran and Xiong, Caiming and Ramaiah, Chetan},
  title     = {Use All the Labels: A Hierarchical Multi-Label Contrastive Learning Framework},
  booktitle = CVPR,
  month     = {June},
  year      = {2022},
  pages     = {16660-16669}
}

@inproceedings{Kim2023gapOfPartialLabel,
  author    = {Kim, Youngwook and Kim, Jae Myung and Jeong, Jieun and Schmid, Cordelia and Akata, Zeynep and Lee, Jungwoo},
  title     = {Bridging the Gap Between Model Explanations in Partially Annotated Multi-Label Classification},
  booktitle = CVPR,
  month     = {June},
  year      = {2023},
  pages     = {3408-3417}
}

@inproceedings{Zhang2023learnLongTailedPartial,
  author    = {Zhang, Wenqiao and Liu, Changshuo and Zeng, Lingze and Ooi, Bengchin and Tang, Siliang and Zhuang, Yueting},
  title     = {Learning in Imperfect Environment: Multi-Label Classification with Long-Tailed Distribution and Partial Labels},
  booktitle = ICCV,
  month     = {October},
  year      = {2023},
  pages     = {1423-1432}
}

@misc{yuan2024positivelabelneedmultilabel,
      title={Positive Label Is All You Need for Multi-Label Classification}, 
      author={Zhixiang Yuan and Kaixin Zhang and Tao Huang},
      year={2024},
      eprint={2306.16016},
      archivePrefix={arXiv},
      primaryClass={cs.CV},
      url={https://arxiv.org/abs/2306.16016}, 
}

@article{hu2023dualcoopfasteffectiveadaptation,
  title={Dualcoop++: Fast and effective adaptation to multi-label recognition with limited annotations},
  author={Hu, Ping and Sun, Ximeng and Sclaroff, Stan and Saenko, Kate},
  journal=PAMI,
  year={2023},
  publisher={IEEE}
}

@article{chen2023semantic,
  title={Semantic Contrastive Bootstrapping for Single-Positive Multi-label Recognition},
  author={Chen, Cheng and Zhao, Yifan and Li, Jia},
  journal=IJCV,
  volume={131},
  number={12},
  pages={3289--3306},
  year={2023},
  publisher={Springer}
}

@misc{xie2021detco,
      title={DetCo: Unsupervised Contrastive Learning for Object Detection}, 
      author={Enze Xie and Jian Ding and Wenhai Wang and Xiaohang Zhan and Hang Xu and Zhenguo Li and Ping Luo},
      year={2021},
      eprint={2102.04803},
      archivePrefix={arXiv},
      primaryClass={cs.CV}
}

@ARTICLE{dai2022updetr,
  author={Dai, Zhigang and Cai, Bolun and Lin, Yugeng and Chen, Junying},
  journal=PAMI, 
  title={Unsupervised Pre-Training for Detection Transformers}, 
  year={2022},
  volume={},
  number={},
  pages={1-11},
  doi={10.1109/TPAMI.2022.3216514}}

@inproceedings{xie2021unsupervised,
  title={Unsupervised Object-Level Representation Learning from Scene Images},
  author={Xie, Jiahao and Zhan, Xiaohang and Liu, Ziwei and Ong, Yew Soon and Loy, Chen Change},
  booktitle=NIPS,
  year={2021}
}

@article{krishna2017visual,
  title={Visual genome: Connecting language and vision using crowdsourced dense image annotations},
  author={Krishna, Ranjay and Zhu, Yuke and Groth, Oliver and Johnson, Justin and Hata, Kenji and Kravitz, Joshua and Chen, Stephanie and Kalantidis, Yannis and Li, Li-Jia and Shamma, David A and others},
  journal=IJCV,
  volume={123},
  pages={32--73},
  year={2017},
  publisher={Springer}
}

@article{chen2024heterogeneous,
  title={Heterogeneous semantic transfer for multi-label recognition with partial labels},
  author={Chen, Tianshui and Pu, Tao and Liu, Lingbo and Shi, Yukai and Yang, Zhijing and Lin, Liang},
  journal=IJCV,
  volume={132},
  number={12},
  pages={6091--6106},
  year={2024},
  publisher={Springer}
}

@inproceedings{cubuk2020randaugment,
  title={Randaugment: Practical automated data augmentation with a reduced search space},
  author={Cubuk, Ekin D and Zoph, Barret and Shlens, Jonathon and Le, Quoc V},
  booktitle={Proceedings of the IEEE/CVF conference on computer vision and pattern recognition workshops},
  pages={702--703},
  year={2020}
}

@InProceedings{Alfassy2019laso,
author = {Alfassy, Amit and Karlinsky, Leonid and Aides, Amit and Shtok, Joseph and Harary, Sivan and Feris, Rogerio and Giryes, Raja and Bronstein, Alex M.},
title = {LaSO: Label-Set Operations Networks for Multi-Label Few-Shot Learning},
booktitle = {Proceedings of the IEEE/CVF Conference on Computer Vision and Pattern Recognition (CVPR)},
month = {June},
year = {2019}
}

@InProceedings{Simon2022mlfsl,
    author    = {Simon, Christian and Koniusz, Piotr and Harandi, Mehrtash},
    title     = {Meta-Learning for Multi-Label Few-Shot Classification},
    booktitle = {Proceedings of the IEEE/CVF Winter Conference on Applications of Computer Vision (WACV)},
    month     = {January},
    year      = {2022},
    pages     = {3951-3960}
}

@misc{zhang2021tipadaptertrainingfreeclipadapterbetter,
      title={Tip-Adapter: Training-free CLIP-Adapter for Better Vision-Language Modeling}, 
      author={Renrui Zhang and Rongyao Fang and Wei Zhang and Peng Gao and Kunchang Li and Jifeng Dai and Yu Qiao and Hongsheng Li},
      year={2021},
      eprint={2111.03930},
      archivePrefix={arXiv},
      primaryClass={cs.CV},
      url={https://arxiv.org/abs/2111.03930}, 
}

\balance

\end{document}